\documentclass{article}

\usepackage[nonatbib, preprint]{neurips_2021} 

\usepackage{multirow}
\usepackage{setspace}
\usepackage[bottom]{footmisc}

\usepackage{graphicx}
\usepackage{booktabs}
\usepackage{makecell}
\usepackage{array}
\usepackage{bm}
\usepackage{balance}
\usepackage{amsmath}
\usepackage{colortbl}
\usepackage{breqn}
\usepackage{tcolorbox}
\usepackage{setspace}
\usepackage{soul}
\usepackage[export]{adjustbox}

\usepackage[utf8]{inputenc} 
\usepackage[T1]{fontenc}    
\usepackage{hyperref}       
\usepackage{url}            
\usepackage{booktabs}       

\usepackage{amsfonts}       
\usepackage{nicefrac}       
\usepackage{microtype}      
\usepackage{graphicx} 
\usepackage{subfig}
\usepackage{wrapfig}
\usepackage{multirow}
\usepackage[T1]{fontenc}
\usepackage{lmodern}

\begin{document}

\title{Fairness-aware Summarization for Justified Decision-Making}

\author{%
  \hspace{-1.5cm} Moniba Keymanesh\\
 \And
Tanya Berger-Wolf\\
\And
Micha Elsner\\
\And
 Srinivasan Parthasarathy\\ 
\And 
{\normalfont The Ohio State University}\\
\texttt{\{keymanesh.1, berger-wolf.1, elsner.14, parthasarathy.2\}@osu.edu} \\
   }
\maketitle

\begin{abstract}

In consequential domains such as recidivism prediction, facility inspection, and benefit assignment, it's important for individuals to know the decision-relevant information for the model's prediction. In addition, predictions should be fair both in terms of the outcome and the justification of the outcome. In other words, decision-relevant features should provide sufficient information for the predicted outcome and should be independent of the membership of individuals in protected groups such as race and gender. In this work, we focus on the problem of (un)fairness in the justification of the text-based neural models. We tie the explanatory power of the model to fairness in the outcome and propose a fairness-aware summarization mechanism to detect and counteract the bias in such models.  Given a potentially biased natural language explanation for a decision, we use a multi-task neural model and an attribution mechanism based on integrated gradients to extract high-utility and low-bias justifications in form of a summary.  The extracted summary is then used for training a model to make decisions for individuals. Results on several real world datasets suggest that our method drastically limits the demographic leakage in the input (fairness in justification) while moderately enhancing the fairness in the outcome. Our model is also effective in detecting and counteracting several types of data poisoning attacks that synthesize race-coded reasoning or irrelevant justifications. 

\end{abstract}


\maketitle

\section{Introduction}

AI systems are increasingly adopted to assist or replace humans in several highly consequential domains including recidivism assessment~\cite{barry-jester_casselman_goldstein_2015}, policing~\cite{rudin2013predictive, keymanesh2020twitter}, credit card offering~\cite{steel_web_cutting}, lending~\cite{koren_search}, and prioritizing resources for inspection~\footnote{https://chicago.github.io/food-inspections-evaluation/}. To maximize the utility, such models are trained to minimize the error on historical data (decisions made by humans in the past). However, the historical decisions can have unfair outcomes or be based on unfair arguments. Training models on historical decisions with unfair outcomes or justifications can reinforce the biases that already exist in our society.  In fact, training models without fairness considerations has already resulted in several cases of discrimination~\cite{kleinberg2016inherent,osoba2017intelligence, angwin2016machine, bolukbasi2016man}. Discrimination in this context is defined as the unjustified distinction between individuals based on their membership in a protected group (e.g. gender identity or ethnicity). The concerns and observations regarding the unfairness of AI algorithms have led to a growing interest in defining, measuring, and mitigating algorithmic unfairness~\cite{pessach2020algorithmic,berk2018fairness,chouldechova2018frontiers, friedler2019comparative, holstein2019improving}. A large body of research on fairness of AI has focused on mitigating the bias in decision-making by minimizing the difference between treatment and outcome among different protected groups (see \S~\ref{related_work}).
\begin{wrapfigure}{l}{0.4\linewidth}
\centering
\includegraphics[width=0.65\linewidth]{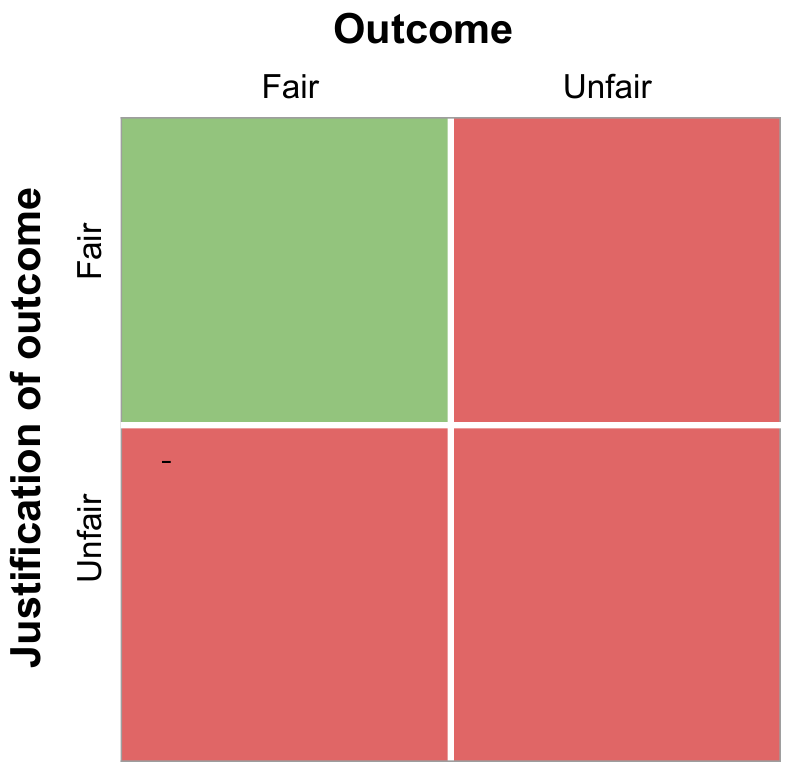}
\caption{A fairly-justified decision should have a fair outcome and be based on fair justifications. } 
\label{fig:fairness_dimensions}
\end{wrapfigure}
 
While training models on historical decisions with unfair outcomes is detrimental,  using historical training data with unfair justifications is equally harmful. For example, training a text-based neural model on unfair justifications can cause the model to associate a gender or race-coded phrase in the input to a certain outcome. This phenomena is an example of disparate impact~\cite{barocas2016big, zafar2017fairness}. On the other hand, it is possible that individuals from two or more protected groups are apparently treated differently (received different outcomes). But the differences can be justified and explained using some fair arguments and therefore is not considered illegal~\cite{mehrabi2019survey}. For example, Kamiral et al ~\cite{kamiran2013explainable} state that the difference in income level in females and males in the UCI adult income dataset~\footnote{https://archive.ics.uci.edu/ml/datasets/adult} --- a well-studied dataset in algorithmic fairness research --- can be attributed to the difference in working hours. Methods that do not take into account the explainability aspect of discrimination will result in reverse discrimination~\cite{kamiran2013explainable}.  This highlights the need to distinguish between the fairness of outcome and fairness in justification of the outcome (see Figure~\ref{fig:fairness_dimensions}). A fairly-justified decision should both have a fair outcome and be fairly justified. In other words, the justification should include enough information to explain the outcome~\cite{carvalho2019machine} and should not be based on information about membership in protected groups.
While there are several sources for unfairness in reasoning of AI models, in this work, we focus on detecting and counteracting biases in the justification of text-based decision-making models. We propose a fairness-aware summarization mechanism as a pre-processing step to reduce potential biases from textual justifications.  We propose methods to first identify and measure bias in textual explanations and then mitigate this bias using a filtering-based approach. We measure bias by using metrics such as demographic parity~\cite{calders2009building}, equalized odds~\cite{hardt2016equality}, and calibration~\cite{kleinberg2016inherent}, and  by measuring the adversary's ability to identify membership in protected groups given the textual explanations. To counteract the bias, our proposed  summarization model obfuscates the arguments that are not useful for decision making or are only useful when they correlate with the protected attribute. Finally, the extracted fairly-justified summaries are used to train a final model. This preprocessing approach ensures learning a model that is both transparent and agnostic about gender-coded or race-coded arguments~\footnote{Note that we do not claim or assume that “fair explanations should avoid mentioning the protected attribute”. Rather methodologically, we are removing the signals about the protected attribute to test whether the rest of the arguments still sufficiently justify the outcome.}. Our framework can potentially assist users in understanding the decisions that are made for them by presenting the most predictive justifications. To summarize, in this study, we make the following contributions: 

\begin{itemize}

\item  We propose the use of a multi-task model and an attribution mechanism to attribute the decision of the model as well as potential biases in the justification to certain parts of the inputs. 

\item We propose a fairness-aware summarization model to condense the input explanations that extracts the decision-relevant justification while removing the potentially unfair ones. Our proposed preprocessing approach is independent of modeling and can be integrated in a data science pipeline with other in-processing and post-processing fairness enhancement mechanisms. 

\item  We show that this pre-processing step does not hurt the utility of the model but significantly limits the leakage of information about protected attributes of individuals in the input justifications.

\item  We show that using our proposed approach to obfuscate the race-coded or gender-coded input justifications moderately enhances the fairness in the outcome.

\item  We test the performance of our proposed approach under several types of unfairness attacks.     

\end{itemize}
Next, we will formally define our problem and explain our proposed solution.


\section{Related Work}
\label{related_work}
 
\textbf{Machine Learning Fairness: }Techniques proposed to enhance fairness in machine learning algorithms can be broadly categorized into pre-processing methods, in-processing methods, and post-processing methods~\cite{pessach2020algorithmic}. Pre-processing mechanisms use re-weighting, relabeling or other transformations of the input data to remove dependencies between the class label and the sensitive attributes before feeding it to the machine learning algorithm~\cite{ghassami2018fairness,calmon2017optimized,zemel2013learning,feldman2015certifying, del2018obtaining,edwards2015censoring,xu2018fairgan}. This class of approaches are closely related to the field of privacy~\cite{ebrahimi2020mobile}. Since both fairness and privacy can be enhanced by obfuscating sensitive information from the input data with the adversary goal of minimal data perturbation~\cite{kazemi2018scalable,jaiswal2020privacy}. In-processing methods modify the optimization procedure of the classifier to integrate fairness criteria in the objective function~\cite{kamishima2012fairness, aghaei2019learning, calders2010three}. This is often done by using a regularization term~\cite{donini2018empirical,zafar2017fairness, zafar2017parity,zafar2017fairness2,goel2018non, bechavod2017penalizing,berk2017convex, rahmattalabi2020fair,kamiran2010discrimination}, meta-learning algorithms~\cite{celis2019classification}, reduction-based methods~\cite{agarwal2018reductions, cotter2019training}, or adversarial training~\cite{madras2018learning, zhang2018mitigating, celis2019improved, wadsworth2018achieving}. Post-processing methods adjust the output of the AI algorithm to enhance fairness in decisions~\cite{fish2016confidence}. For example, by flipping some of the decisions of the classifier~\cite{hardt2016equality} or learning a different classifier~\cite{dwork2018decoupled} or a separate threshold for each group~\cite{menon2018cost}. Our proposed approach of using fairness-aware text summarization to remove bias from the input explanations belongs to the first category. Majority of the introduced methods mitigate bias in decision-making by minimizing the difference between treatment and outcome among different protected groups. Our proposed approach is distinct with previous work in a few ways. In contract to the approaches that are intended to enhance the fairness of the model's outcome, our proposed approach is intended to enhance the fairness in the justification of the outcome. Moreover, many of the existing preprocessing approaches produce an intermediate data representation which is not interpretable to many stakeholders~\cite{ghassami2018fairness, zemel2013learning}. The output of our model is an extractive summary of the input justifications. This is preferable for many applications where interpretability is essential. 

\vspace{0.3cm}
\textbf{Text Summarization: }Our work is also related to the field of automatic text summarization. The general goal of this task is to shorten a text while preserving the key information. Automatic summarization methods can be broadly categorized as abstractive~\cite{rush2015neural,see2017get,lin2019abstractive} and extractive~\cite{nallapati2016classify,sarkhel2020interpretable}. Our work belongs to the latter category.  In extractive summarization, a subset of phrases or sentences in the input document are selected based on an importance score to be included in the final summary.  Defining importance is highly domain specific. However, for extracting generic summaries earlier work has explored using heuristics such as frequency of significant words, coverage of salient concepts~\cite{ dorr2003hedge,filippova2013overcoming}, or the centrality in the document graph~\cite{mihalcea2004textrank} to rank and select sentences in a document. More recently, data-driven approaches rely on deep neural models to extract summaries by creating sentence representation and training a supervised model to learn whether to include a sentence in the summary or not~\cite{kryscinski2019neural,liu2019fine}. While extractive summarization has proved a great solution for applications such as   privacy~\cite{manor-li-2019-plain,keymaneshtoward,keymanesh2021privacy}, and legal decision making~\cite{farzindar2004legal,kanapala2019text,zhong2019automatic}, to the best of our knowledge we are the first to use text summarization for detecting and obfuscating biases in the input data while preserving the decision-relevant information.



\section{Problem Formulation}
Given a dataset consisting of $n$ samples $\{(X_i,Y_i,P_i)\}^n_{i=1}$ where $X$ denotes a textual explanation written by the decision-maker to provide evidence or justify an outcome $Y$ and $P$ indicates one or more protected variables\footnote{We only assume the existence of a set of discrete predefined protected attributes that are relevant to the problem in hand. An example of this is protected groups in the US legal system such as race, gender, and nationality. However, the proposed approach is intended to work with any set of predefined groups. }, we aim to extract a fairly-justified summary $\{{X_i}'\}^n_{i=0}$ such that $X'$ provides sufficient information to predict and justify $Y$ and $X'$ is independent of protected variable $P$. We explain how we measure and attribute these qualities to sentences in the justification $X$ in \S~\ref{s:proposed_methodology}. For instance, $Y_i$ could represent a court decision for individual $i$, which is a member of the demographic group $P_i$ and has received a textual argument $X_i$ regarding this decision\footnote{While we assume availability of information about individual's protected groups at train time, we do not assume that $P_i$ is known at inference time.}. Potentially, $X_i$ can be biased toward certain demographic groups. Our goal is to transform a given dataset $\{(X_i,Y_i,P_i)\}^n_{i=1}$ into a new dataset  $\{({X_i}',Y_i,P_i)\}^n_{i=1}$ that is decontaminated from unfair arguments. To achieve this goal, we use a  fairness-aware extractive summarization model as a data pre-processing step. 

\section{Proposed Methodology}
\label{s:proposed_methodology}

\begin{wrapfigure}{r}{0.42\linewidth}
\centering
\includegraphics[width=0.70\linewidth]{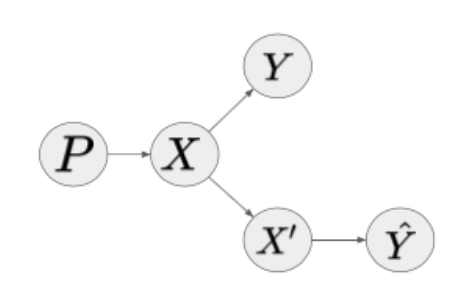}
\caption{A graphical model of the proposed approach. $P$ represents the protected attribute. $X$ indicates the input explanations while $X'$ indicates the farily-justified summary of $X$ which is used to train the final model to predict outcome $\hat{Y}$.}  
\label{fig:graphical_model_summarization}
\end{wrapfigure}

In this section, we explain our proposed methodology to extract a fairly-justified summary $\{{X_i}'\}^n_{i=0}$ such that summary $X'$ provides sufficient information to predict and justify $\hat{Y}$ and the extracted summary $X'$ is independent of protected variable $P$.  A graphical model of the proposed approach is shown in Figure~\ref{fig:graphical_model_summarization}.  Given an input explanation $X_i$ consisting of sentences $\{s_1, s_2, ..., s_m\}$, the goal of our model is to select a subset of these sentences subject to a utility and a fairness constraint. Next,  we explain how we measure and attribute utility and discrimination of the input sentences.  

\textbf{Utility Control: } 
\label{s:utility_cotrol}
To ensure that the extracted summary $X'$ includes sufficient decision-relevant information in $X$, we measure the salience of each sentence in $X$ in predicting outcome $Y$. We train a neural classification model on $X$ using ground truth decision $Y$ as supervision. Next, we use this model to derive the contribution of each sentence in $X$ for predicting outcome $\hat{Y}$. This process is explained in \S~\ref{s:attribution}. To ensure learning generalizable patterns, we hypothesize that the dataset is sufficiently large and the model can learn which factors are associated with which outcomes. This assumption especially holds for scenarios in which a decision-maker (e.g. an inspector or judge) is required to go through a standard set of criteria (e.g. a standard form or set of guidelines) and thus, the same arguments may repeatedly be articulated in different ways to justify a certain outcome.


\textbf{Discrimination Control: }
\label{s:discrimination_control}
To ensure that sentences in input explanation $X$ that are biased toward certain protected groups are excluded from summary $X'$, we attribute a discrimination score to each sentence in $X$. To do so, we measure the utility of an argument in identifying the membership of an individual $i$ in the protected group $P_i$. Note that, we do not assume that “if you mention the protected attribute in the justification you are being unfair”. Rather, methodologically, we remove the signals about the protected attribute to test whether the rest of the arguments in $X'$ still sufficiently justify the outcome. Moreover, this is a way of demonstrating to the stakeholders that the model decision is not conditioned on the protected attribute. If removing the gender or race-coded language from the justifications does not change the predicted outcome, then we can conclude that the initial gender or race-coded language (that was removed) was not an unfair justification. To measure the discrimination score,  we use justification $X$ to predict protected attribute $P$. Next, we use the trained model to derive the contribution of each sentence in the membership identification task. Sentences with a high discrimination score are removed.  We train a multi-task model for decision classification and membership identification tasks. This process is explained in the next Section. 
\vspace{1cm}

\subsection{Model Architecture}
\label{s:neural_text_classification}

\begin{wrapfigure}{t}{0.5\linewidth}
\begin{center}

\includegraphics[width=0.85\linewidth]{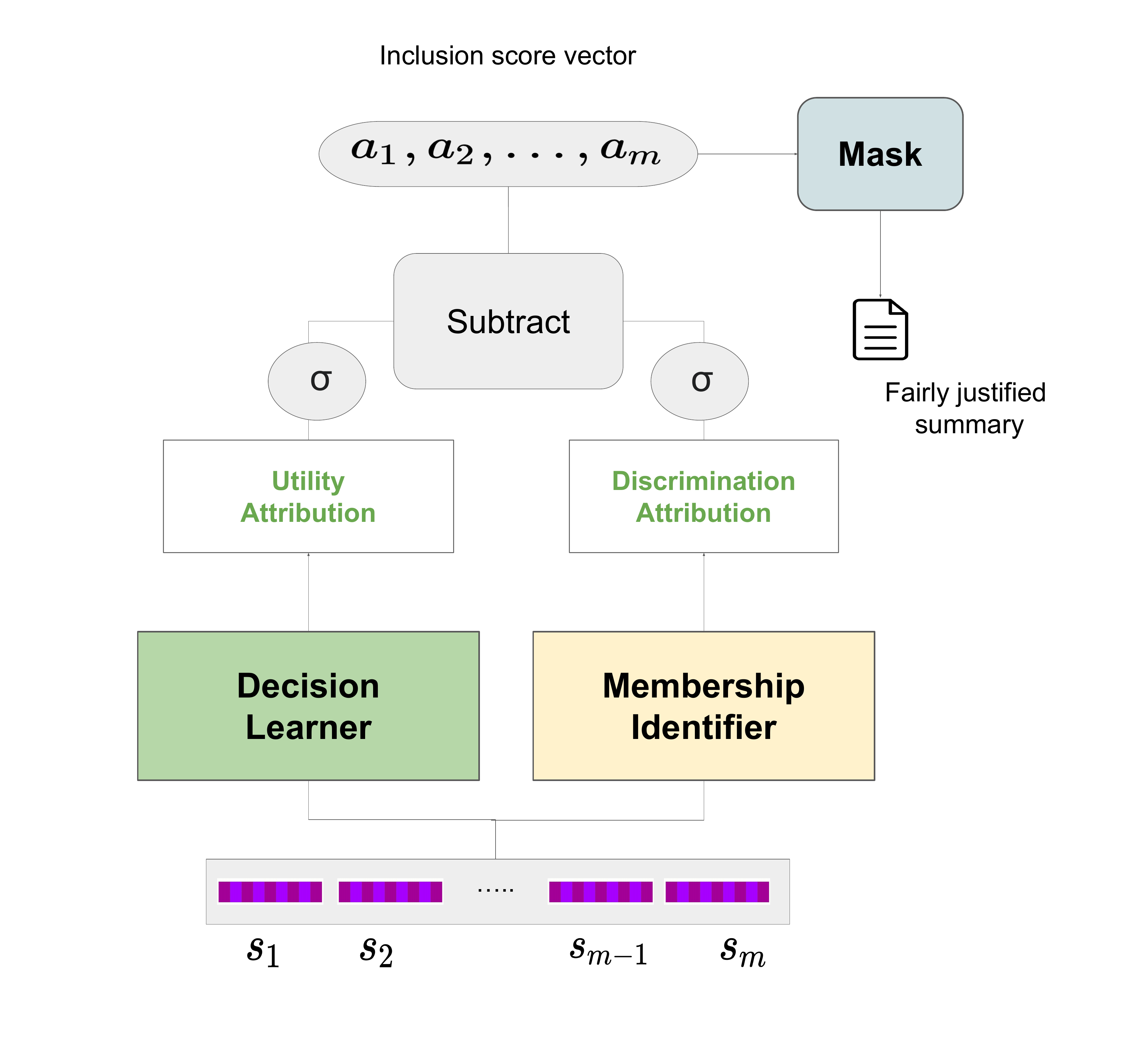}
\end{center}
\begin{small}
 \vspace{-0.5cm}
 \caption{An overview of the architecture: Decision learner and membership identifier are trained using decision $Y$ and protected attribute $P$ as supervision respectively. The attributions of each module is normalized and subtracted to obtain the inclusion scores.} 
 \label{fig:attribution_arc}
\end{small}

\end{wrapfigure}

Prior research has adopted word embeddings and Convolutional Neural Networks~(CNN) for variety of sentence classification tasks~\cite{collobert2011natural, kalchbrenner2014convolutional,heidari2020using, jafariakinabad2019style,zhang2015sensitivity,heidari2020semantic}. Kim~\cite{kim2014convolutional} achieved strong empirical performance using static vectors and little hyper-parameter tuning over a range of benchmarks.  Variations of this architecture have achieved good performance for extractive summarization of privacy policies~\cite{keymaneshtoward}
and court cases~\cite{zhong2019automatic}. CNNs are fast to train and can easily be combined with methods such as Integrated Gradients~\cite{sundararajan2017axiomatic} for attributing predictions to specific parts of the input. These considerations led to our decision to use a slight variant of the sentence-ngram CNN model in~\cite{zhong2019automatic} for decision outcome prediction and membership identification tasks. Given explanation $X_i$ consisting of $m$ sentences/arguments $\{s_1, ....s_m\}$ to justify decision $Y_i$ for individual $i$, we use Universal Sentence Encoder~\cite{cer2018universal} to encode each sentence $s_j$ to a 512-dimensional embedding vector $v_j$. We build the justification matrix $A \in R^{m \times 512}$ by concatenating the sentence vectors $v_1$ to $v_m$:
\[
 A_{1:m} =  v_1 \oplus v_2 \oplus ... v_m
\]
The Sentence Encoder is pre-trained using a variety of data sources and tasks~\cite{cer2018universal} using the Transformer~\cite{vaswani2017attention} architecture
and is obtained from \href{https://tfhub.dev/google/universal-sentence-encoder/4}{Tensorflow Hub }. Following~\cite{collobert2011natural} we apply convolution filters to windows of sentences in explanation $X_i$ to capture compounded and higher-order features. We use multiple filter sizes to capture various features from sentence n-grams. We use filter sizes of $h \times d$ where $h$ is the height or region size of the filter and indicates the number of sentences that are considered jointly when applying the convolution filter. $d$ is the dimensionality of the sentence vectors and is equal to 512. The feature map $c \in R^{m-h+1}$ of the convolution operation is then obtained by repeatedly applying the convolution filter $w$ to  a window of sentences $s_{j:j+h-1}$. Each element $c_j$ in feature map $c = [c_1,c_2, ...c_{m-h+1}]$ is then obtained from:
\[
 c_i = f( w~.~A[j:j+h-1] + b)
\] 
where $A[j:k]$ is the sub-matrix of $A$ from row $j$ to $k$ corresponding to  a window of sentence $s_{j}$ to $s_k$ and "$.$" represents the dot product between the filter $w$ and the sub-matrices. $b \in R$ represents the bias term and $f$ is an activation function such as a rectified linear unit. We use window sizes 2, 3, and 4 and train 100 filters for each window size. The dimensionality of the feature map $c$ generated by each convolution filter is different for explanations with various lengths and filters with different heights.  We apply an average-max pooling operation over the feature maps of each window size to downsample them. Next, we concatenate the output vectors. Eventually, the concatenated vector runs through a dense layer with 64 units followed by an activation function\footnote{For classification tasks we used softmax (multi-class) or Sigmoid (binary classes) functions. For scalar outputs, we used Rectified Linear Unit.}.  This is a multi-task model with a decision learner and membership identifier modules. The decision learner is trained using decision outcome $Y$ as supervision and the membership identifier is trained using the protected attribute $P$.  The loss at each epoch is computed based on a weighted sum of the decision prediction and membership identification losses. Training details are explained in Appendix~\ref{fairness-training-details}. Next, we explain the method we use for attributing the predictions $\hat{Y}$ and $\hat{P}$ of the model to arguments in $X$.  

\subsection{Attribution}
\label{s:attribution}

Sundararajan et al~\cite{sundararajan2017axiomatic} proposed a method called Integrated Gradients to attribute predictions of a deep neural network to its input features. This method is independent of the specific neural architecture and can provide a measure of relevance for each feature by quantifying its impact on the predicted outcome. Zhong et al~\cite{zhong2019automatic} adopted this method for identifying most decision-relevant aspects of legal cases. 
We also utilize this method to measure the impact of each input sentence in decision prediction and membership identification tasks.
Essentially we take a straight line path from input $x$ to its baseline $b$ \footnote{Conceptually, baselines represent data points that do not contain any useful information for the model. They are used as a benchmark by the integrated gradients method. Sundararajan et al~\cite{sundararajan2017axiomatic} suggest using an all-zero input embedding vector for text-based networks.} and notice how model prediction changes along this path by integrating the gradients along the path. To approximate the integral of the integrated gradients,  we simply sum up the gradients at points occurring at small intervals along the straight-line path from the baseline to the input. The resulting single scalar represents the gradients and attributes the prediction to input features. The integrated gradient along the $i$-th dimension for an input $x$ and baseline $b$ is defined as follows: 
\[
 IG_i(x) ::= (x_i - b_i) \times \sum_{k=1}^{m} \frac{ \partial{F(b+ \frac{k}{m} \times (x-b) )}} {  \partial x_i } \times \frac{1}{m}  
\] 

Here, $F: X \rightarrow Y$ represents the neural model, $\frac{\partial F(x)}{\partial x_i}$ is the gradient of F(X) along the $i$-th dimension, $x$ represents the input at hand, $b$ represents the baseline input (an all-zero vector), and  $m$ is the number of steps in the approximation of the integral\footnote{Sundararajan et al~\cite{sundararajan2017axiomatic} applied Integrated Gradients to a variety of deep architectures including CNN. The only assumption that they make is that function F should be differentiable almost everywhere. Deep networks built out of Sigmoids, ReLUs, and pooling operators satisfy this condition.}. 
To obtain utility attribution $U = \{u_1, u_2, ...u_m\}$ for sentences $\{s_1, s_2, ..., s_m\}$  in input justification $X_i$ we calculate the attributions for the model using the predicted decision outcome $\hat{Y}$. Note that each input feature is one dimension of sentence embedding. To obtain salience scores for each sentence, we sum up the attribution scores for each dimension. Next, we run $U$ through a softmax function to get a utility distribution over the sentences. Similarly, we obtain discrimination attribution $D = \{d_1, d_2, ...d_m\}$ for  sentences $\{s_1, s_2, ..., s_m\}$  by calculating the integrated gradients attributions for the model using the predicted protected attribute $\hat{P}$. We run $D$ through a softmax function to get a discrimination distribution over the sentences. We include high-utility and low-bias sentences in the fairly-justified summary of the explanations. The final inclusion score $a_i$ for each sentence is computed using the following equation\footnote{Both U and D satisfy properties of a probability distribution as $\Sigma^{m}_{i=1} u_i = 1$ and $ 0 \leq u_i  \leq 1$. Thus, each $u_i$ and $d_i$ have comparable scales}: 
\[
 a_i = \sigma(u)_i -  \alpha \times \sigma(d)_i
\] 
In the equation above, $\alpha$ is a hyper-parameter that controls the utility-discrimination trade-off. Higher values of $\alpha$ correspond to removing more information about protected attribute from the input justification. Figure~\ref{fig:attribution_arc} shows the attribution process. Methodologically, we want to identify and remove arguments that are not useful for decision prediction or are only useful for prediction of outcome when they are also helping in prediction of the protected attribute. The subtraction operation ensures that such arguments get a small inclusion score $a_i$.

\paragraph{\textbf{Extracting Fairly-Justified Summarizes: }}
\label{s:extracting-fairly-justified-summaries}
Given sentences $\{s_1, s_2, ..., s_m\}$ and the corresponding inclusion scores $\{a_1, ....a_m\}$, we select sentences with a positive score for inclusion in the output summary. These sentences have high utility for decision prediction but do not reveal the protected attribute of the individuals. We refer to our preprocessing method as \textit{FairSum}. In our experiments, we test whether training a decision classifier on justifications pre-processed by FairSum will enhance fairness in the justification on real-world and synthetic datasets.

\section{Experiments and Results}
In this section, we introduce the datasets we use for training and testing our model followed by  experimental setup and metrics in consideration. 

\subsection{Datasets}
\label{fairly-justified-dataset}
\textbf{Inspection Reports of food establishments in Chicago (D1):} The City of Chicago has published reports of food inspections conducted since 2010. We extracted the information on food inspections conducted from January 2010 till December 2014 from the City of Chicago's GitHub repository\footnote{https://github.com/Chicago/food-inspections-evaluation}. This dataset contains the outcome of inspection which can be \textit{pass}, \textit{fail}, or \textit{conditional pass} as well as notes that the sanitarian left in the inspection form about the observed violations in order to justify the outcome and explain what needs to be fixed before the next visit\footnote{There could be other outcomes e.g. when the sanitarian could not access the establishment. These cases are excluded from our study.}. In food inspections, decision are being made for both the restaurant owner and the public health. In this work, we focus on fairness with respect to customers of the food establishment. Thus, we consider the ethnicity of the majority of the population in the census block group that the food establishment is located at as the protected attribute. This is a reasonable proxy given that Chicago is one of the most segregated cities in the US~\cite{chicago-segregated}\footnote{The demographic information of neighborhoods were extracted from https://www.census.gov/}.  This dataset includes 17,212 inspection reports. The inspector comments are on average 18.2 sentences long with a standard deviation of 7.2. The breakdown of the inspection outcome for each demographic group is shown in Table~\ref{inspection-results-table} of Appendix~\ref{appx:dataset-stats}.  We train the model explained in \S~\ref{s:neural_text_classification} on inspector notes using inspection outcome and the ethnicity of the majority of the customers as supervision for decision classifier and membership identifier respectively. We use 90\% of inspections from January 2010 till October 2013 (75\% all records in our data-set) as our training set and the remaining 10\% as our validation set. The inspections conducted from November 2013 till December 2014 are used as our test set. We represent this dataset with D1.

\textbf{Rate My Professor (D2-D4): }Students can leave an anonymous review and rating on the scale of 1-5 in several categories for their instructors on the ~\href{https://www.ratemyprofessors.com/}{Rate My Professor (RMP)} website. Previous work has identified several types of biases in students' evaluations~\cite{legg2012ratemyprofessors, reid2010role,clayson2014does, bleske2010ratemyprofessors, rosen2018correlations, theyson2015hot}. In our study, we aim to detect and remove potential biases in justifications provided by students to explain their ratings. 
We rely on the dataset collected by He et al~\cite{https://doi.org/10.17632/fvtfjyvw7d.2}. We combine all the reviews written for each instructor and use the average rating as the supervision for the decision classifier. We use the gender of the instructor as the supervision for the membership identifier model. In our experiments, we exclude the instructors that have less than 5 reviews. We also remove the pronouns and instructors' names from the reviews.\footnote{This pre-processing step ensures that the membership identifier does not rely on blatant signals from the text and instead extracts more latent patterns in the justifications.} The resulting dataset includes reviews written for 1344 instructors which are on average 45.6 sentences long. We indicate this dataset with D2.

Prior work, has shown that using gender-coded language in teaching evaluations is more common in disciplines with a large gender-gap~\cite{storage2016frequency}. Inspired by this observation and to study the impact of reviewer's gender (students) on teaching evaluations, we create two additional datasets D3 and D4. To do so, we split the RMP dataset based on the gender gap of the students in each discipline. D3 includes student evaluations for professors in fields that are female-dominant such as nursing, psychology, and education while D4 includes student evaluations for male-dominant majors such as engineering, computer science, and philosophy~\footnote{Fields with less than 20\% gender gap are excluded. The statistics about the bachelor’s degrees earned by field and gender is obtained from~\cite{degreebygender}  }. For D2-D4, we randomly split our dataset to a 70-15-15 split to build our train, validation, and test sets. The breakdown of ratings for each gender group for D2-D4 is shown in  Appendix~\ref{appx:dataset-stats}. Our training details and hyper parameter setup can be found in Appendix~\ref{fairness-training-details}. 
\subsection{Evaluation Metrics}
The current automatic evaluation protocol for automatic text summarization is based on similarity of the model-generated summary to a human-written summary and using metric such as ROUGE~\cite{lin2004rouge}.  
Shandilya et al~\cite{shandilya2018fairness} was the first to evaluate text summarization systems from fairness perspective. They verify fairness of summaries using the notion of "adverse-impact" by measuring the fraction of selected tweets to be incorporated in the output summary from each protected group. The goal of our work (enhancing fairness in justification while preserving decision-relevant information) however, is different from traditional text summarization as well as the notion of fairness used in~\cite{shandilya2018fairness}. Thus,  we use a new perspective for evaluation of extracted summarizationa which is based on demographic leakage and fairness of outcome.  Essentially, in our experiments we seek to answer the following questions:  (i) \textit{How does applying FairSum on the input justification impact the utility of the model?}  (ii) \textit{Will this pre-processing step effectively remove the proxy information about the protected attribute from the justifications?} and (iii) \textit{How will this preprocessing approach impacts the fairness of the outcome?} (vi) \textit{Is FairSum able to mitigate different types of unfairness attacks?} 

To answer the first question, we report the utility of the decision learner. For categorical outcomes (e.g. in D1) we report the Micro-F1 and Macro-F1 and for scalar outcomes (D2-D4) we report the Mean Absolute Error(MAE). To answer the second question, we report the demographic leakage. Leakage is defined as the ability of the membership identifier network to correctly predict the protected attribute of the individuals given the justification. We report the Micro-F1 and Macro-F1 of our membership identification model. Lower demographic leakage is desirable.

While FairSum is not directly designed to address the fairness of the outcome, we seek to study how enhancing fairness in justification impacts fairness of outcome. To do so,  for categorical outcomes we report the demographic parity, equality of odds, and calibration. For each of these metrics we report the gap between the most favored and least favored group. For a discussion on fairness measures and their trade-offs see~\cite{kleinberg2016inherent} and~\cite{hardt2016equality}. We additionally report False Pass Rate Gap (FPRG) and False Fail Rate Gap (FFRG) across demographic groups. FPRG and FFRG represent the equality in distribution of the model errors across demographic groups. Similar metrics were used in~\cite{wadsworth2018achieving}. To measure fairness for scalar outcomes (D2-D4), we report the Mean Absolute Error GAP between the demographic groups (male and female).  See Appendix~\ref{app:evaluation metrics} for formal definition of fairness metrics in the context of food inspection. To answer the last question, we perturb dataset D1 to create several types of injection attacks. We measure the attack success before and after applying FairSum on the perturbed data. Our findings are shared in \S~\ref{attacks_experiments}.

\subsection{Results and Discussion}
\label{fairness-experimental-results}

 In our experiments, we compare the utility, demographic leakage, and fairness of models that are identical in terms of architecture but are trained on different versions of the training data. The model architecture is discussed in \S~\ref{s:neural_text_classification}. Our results on dataset D1 is shared in Table~\ref{tab:foodinspection-results-t1} and Table~\ref{tab:foodinspection-results-t2}. Our Results on datasets D2-D4 is shared in Table~\ref{tab:RMP-results}. 
 In the "Empty" setting, justifications are empty. In the "Full" setting, the model is trained and tested on the original data while in the "FairSum" setting it is trained and tested on justifications summarized by FairSum. We use to empty setting to indicate the lower bound of the demographic leakage. We use the full setting, to measure the bias in the justifications in the input dataset. This setting also acts as our baseline.   
 We apply FairSum on both the train and test sets. The parameter $\alpha$ which controls the trade-off between the utility and the demographic leakage is set to 1.
 
 As it can be seen in Table~\ref{tab:foodinspection-results-t1}, FairSum reduces the demographic leakage on dataset D1 (by 0.06 in Micro-F1 and 0.05 in Macro-F1) while achieving the same level of accuracy on the decision classification task in comparison to the full setting. FairSum also decreases parity by 0.01 while achieving similar results in terms of FFRG and FPRG.

\begin{table*}[t]\resizebox{1\linewidth}{!}{

\begin{tabular}{lrrrrrrrrrrrr}
\midrule
\textbf{Dataset} & \multicolumn{3}{c}{\textbf{\begin{tabular}[c]{@{}c@{}}Utility $\uparrow$ \\ \small (Micro-F1)\end{tabular}}}                 & \multicolumn{3}{c}{\textbf{\begin{tabular}[c]{@{}c@{}}Utility $\uparrow$ \\ \small (Macro-F1)\end{tabular}}}                 & \multicolumn{3}{c}{\textbf{\begin{tabular}[c]{@{}c@{}}Demographic Leakage $\downarrow$\\  \small (Micro-F1)\end{tabular}}}  & \multicolumn{3}{c}{\textbf{\begin{tabular}[c]{@{}c@{}}Demographic Leakage $\downarrow$ \\ \small (Macro-F1)\end{tabular}}}  \\ \midrule
                 & \multicolumn{1}{l}{Empty} & \multicolumn{1}{l}{Full} & \multicolumn{1}{l}{\cellcolor[HTML]{C0C0C0}FairSum} & \multicolumn{1}{l}{Empty} & \multicolumn{1}{l}{Full} & \multicolumn{1}{l}{\cellcolor[HTML]{C0C0C0}FairSum} & \multicolumn{1}{l}{Empty} & \multicolumn{1}{l}{Full} & \multicolumn{1}{l}{\cellcolor[HTML]{C0C0C0}FairSum} & \multicolumn{1}{l}{Empty} & \multicolumn{1}{l}{Full} & \multicolumn{1}{l}{\cellcolor[HTML]{C0C0C0}FairSum} \\
D1               & 0.48                      & 0.83                     & \cellcolor[HTML]{C0C0C0}0.83                        & 0.22                      & 0.83                     & \cellcolor[HTML]{C0C0C0}0.82                        & 0.56                      & 0.58                     & \cellcolor[HTML]{C0C0C0}0.52                        & 0.18                      & 0.38                     & \cellcolor[HTML]{C0C0C0}0.33                        \\ \midrule                 
\end{tabular}}
\caption{Results on datasets D1. "$\uparrow$": higher is better. "$\downarrow$": lower is better. \vspace{-0.5cm}} 

\label{tab:foodinspection-results-t1}
\end{table*}

\begin{table*}[t] \resizebox{1\linewidth}{!}{
\begin{tabular}{lrrrrrrrrrr} \midrule
\multicolumn{1}{c}{\textbf{Dataset}} & \multicolumn{2}{c}{\textbf{Parity $\downarrow$ }}                                            & \multicolumn{2}{c}{\textbf{\begin{tabular}[c]{@{}c@{}}Equality \\ of  Odds $\downarrow$ \end{tabular}}} & \multicolumn{2}{c}{\textbf{Calibration $\downarrow$}}                                       & \multicolumn{2}{c}{\textbf{FPRG $\downarrow$}}                                              & \multicolumn{2}{c}{\textbf{FFRG $\downarrow$}}                                              \\ \midrule
                                   & \multicolumn{1}{l}{Full} & \multicolumn{1}{l}{\cellcolor[HTML]{C0C0C0}FairSum} & \multicolumn{1}{l}{Full}      & \multicolumn{1}{l}{\cellcolor[HTML]{C0C0C0}FairSum}      & \multicolumn{1}{l}{Full} & \multicolumn{1}{l}{\cellcolor[HTML]{C0C0C0}FairSum} & \multicolumn{1}{l}{Full} & \multicolumn{1}{l}{\cellcolor[HTML]{C0C0C0}FairSum} & \multicolumn{1}{l}{Full} & \multicolumn{1}{l}{\cellcolor[HTML]{C0C0C0}FairSum} \\
D1                                  & 0.15                     & \cellcolor[HTML]{C0C0C0}0.14                        & 0.08                          & \cellcolor[HTML]{C0C0C0}0.1                              & 0.05                     & \cellcolor[HTML]{C0C0C0}0.06                        & 0.05                     & \cellcolor[HTML]{C0C0C0}0.05                        & 0.11                     & \cellcolor[HTML]{C0C0C0}0.11                        \\
                                     \midrule                    
\end{tabular}}
\caption{Fairness metrics for datasets D1. "$\downarrow$": lower is better.\vspace{-0.5cm} } 
\label{tab:foodinspection-results-t2}
\end{table*}

\begin{table*}[t] \resizebox{1\linewidth}{!}{
\begin{tabular}{lrrrrrrrrrrrr} \midrule
\multicolumn{1}{c}{\textbf{Dataset}} & \multicolumn{3}{c}{\textbf{MAE $\downarrow$}}                                                                           & \multicolumn{3}{c}{\textbf{\begin{tabular}[c]{@{}c@{}}Demographic Leakage $\downarrow$\\ \small (Micro-F1)\end{tabular}}}  & \multicolumn{3}{c}{\textbf{\begin{tabular}[c]{@{}c@{}}Demographic Leakage $\downarrow$ \\ \small (Macro-F1)\end{tabular}}}   & \multicolumn{3}{c}{\textbf{MAE Gap $\downarrow$}}                                                                       \\ \midrule
                                     & \multicolumn{1}{c}{Empty} & \multicolumn{1}{c}{Full} & \multicolumn{1}{c}{\cellcolor[HTML]{C0C0C0}FairSum} & \multicolumn{1}{c}{Empty} & \multicolumn{1}{c}{Full} & \multicolumn{1}{c}{\cellcolor[HTML]{C0C0C0}FairSum} & \multicolumn{1}{c}{Empty} & \multicolumn{1}{c}{Full} & \multicolumn{1}{c}{\cellcolor[HTML]{C0C0C0}FairSum} & \multicolumn{1}{c}{Empty} & \multicolumn{1}{c}{Full} & \multicolumn{1}{c}{\cellcolor[HTML]{C0C0C0}FairSum} \\
D2                                   & 0.72                      & 0.47                     & \cellcolor[HTML]{C0C0C0}0.49                        & 0.59                      & 0.71                     & \cellcolor[HTML]{C0C0C0}0.61                        & 0.37                      & 0.69                     & \cellcolor[HTML]{C0C0C0}0.58                        & 0.07                      & 0.06                     & \cellcolor[HTML]{C0C0C0}0                           \\
D3                                   & 0.76                      & 0.52                     & \cellcolor[HTML]{C0C0C0}0.53                        & 0.5                       & 0.66                     & \cellcolor[HTML]{C0C0C0}0.61                        & 0.33                      & 0.66                     & \cellcolor[HTML]{C0C0C0}0.59                        & 0.19                      & 0.03                     & \cellcolor[HTML]{C0C0C0}0.06                        \\
D4                                   & 0.66                      & 0.54                     & \cellcolor[HTML]{C0C0C0}0.53                        & 0.45                      & 0.82                     & \cellcolor[HTML]{C0C0C0}0.74                        & 0.3                       & 0.71                     & \cellcolor[HTML]{C0C0C0}0.49                        & 0.04                      & 0.02                     & \cellcolor[HTML]{C0C0C0}0     \\ \midrule                     
\end{tabular}}
\caption{Results on RMP Datasets (D2-D4).  "$\downarrow$": lower is better.} 
\label{tab:RMP-results}
\end{table*}

 We see in Table~\ref{tab:RMP-results} that on dataset D2, FairSum decreases the demographic leakage from 0.71 to 0.61 Micro-F1 and 0.69 to 0.58 Macro-F1 while increasing the MAE by 0.02 in a 5-point scale.  FairSum outcomes also are more fair on D2. In the full setting,  predictions have 0.06 higher average MAE for females than males. While FairSum achieves similar error rates for both gender groups (0 MAE gap).

\definecolor{orange}{rgb}{0.93, 0.57, 0.13} 
\definecolor{purple}{rgb}{0.59, 0.44, 0.84} 

\definecolor{ashgrey}{rgb}{0.7, 0.75, 0.71} 

\begin{figure}[t]
\begin{tcolorbox}[colback=white,colframe=black,boxrule=1pt,arc=2pt,boxsep=0pt,left=5pt,right=5pt,top=5pt,bottom=2pt]
\begin{small}

\vspace{0.1cm}
\textbf{Example 1:  } \textcolor{orange}{Not my favorite instructor. We spent a lot of time on things that seemed not important.} \textcolor{purple}{Course syllabus included a lot of topics that have no practical use. Some days the presentations were unclear but I would recommend this course to non-majors. Very open and well organized.} \textcolor{orange}{ The guys in the class love __ .} \textcolor{purple}{__  is a pretty good , but not a great teacher.} \textcolor{orange}{ __ is a great professor also has the physical features that makes you not want to miss a class. Last semester came to class is a short skirt .omg!} \textcolor{orange}{ __ has a lot of experience with undergrad students. Sometimes vague on grading criteria .} \textcolor{purple}{ This is a pretty easy class. -- is very nice.}  \textcolor{orange}{__ is hot and funny.} \textcolor{purple}{ If you get past physical attributes you really learn something. Wow! ... what an interesting topic!} \textcolor{orange}{I respect __ for __ intelligence and ability to teach , not for __ appearance. Very good looking omg!} \textcolor{purple}{ Great course and additional materials are a great support.} 

\end{small}
\end{tcolorbox}
\caption{Applying FairSum on teaching evaluations for a female professor (anonymized and paraphrased for privacy considerations). The pronouns and names have been removed before model training and attribution. Sentences with a positive attribution score (purple) are preserved in the summary $x'$ while the sentences with a negative attribution score (orange) are excluded. \label{fig:example1}\looseness=-1}
\end{figure}


\definecolor{orange}{rgb}{0.93, 0.57, 0.13} 
\definecolor{purple}{rgb}{0.59, 0.44, 0.84} 

\begin{figure}[t]
\begin{tcolorbox}[colback=white,colframe=black,boxrule=1pt,arc=2pt,boxsep=0pt,left=5pt,right=5pt,top=5pt,bottom=2pt]
\begin{small}

\vspace{0.1cm}
\textbf{Example 2:} \textcolor{purple}{Lectures are short. Tests do not really cover what is covered in class. Textbook is not used. Dr. _ is very knowledgeable and passionate about this subject. You will enjoy the class if you are interested in the topic. Highly recommend if you want a nice grade.} \textcolor{orange}{_ is funny , intelligent, and easy to listen to. _  got an epic beard.} \textcolor{purple}{_ post the material online which makes the class very accessible. If you do all the assignments it is impossible to not get an A!} \textcolor{orange}{_is one of very few whom I really think understands the `` real world '' and its workings. I think it is because of __ days in navy. _ is very funny as well.  } \textcolor{purple}{ _ curves quizzes slightly. So, in the end your grade could be better than what you may think. _ can be very helpful , but you must go to the office hours. Probably the easiest five credit class you can take.}

\end{small}
\end{tcolorbox}
\caption{Applying FairSum on teaching evaluations for a male professor (anonymized and paraphrased for privacy considerations). Sentences with a positive attribution score (purple) are preserved in the summary $x'$ while the sentences with a negative attribution score (orange) are excluded. \label{fig:example2}\looseness=-1}
\end{figure}

On D3 and D4, fairSum reduces the demographic leakage (from 0.66 to 0.59 and 0.71 to 0.49 Macro-F1 respectively). FairSum is noticeably effective in removing the gender-coded  language in D4 which is sourced from male-dominated majors with 0.82 gender prediction accuracy in the Full setting. This comes with almost no change in model's utility as the average MSE on D2-D4 for is 0.51 for both Full setting and FairSum. 

We conclude that our proposed approach is very effective in reducing the demographic-leakage in the input justifications while also not reducing the utility of the model. Removing gender-coded language from D3 justifications comes with the cost of having 0.06 higher MAE for females than males (this was 0.03 for the full setting). On D2 and D4 however, FairSum completely closes the MAE gap between the gender groups.  

An example of applying FairSum on a teaching evaluation for a two professors is shown in Figure~\ref{fig:example1} and Figure~\ref{fig:example2}. In Figure~\ref{fig:example1}, we see that arguments about the looks of the instructor (more frequent for female instructors) are excluded from the text (indicated with orange). The preserved sentences are indicated with a purple and have a high inclusion score. In Figure~\ref{fig:example2}, arguments about being "intelligent and funny" (more frequent for male instructors) are removed from $x$ by FairSum. While mentioning "intelligence" is not an unfair argument on it's own, more frequent usage for a certain demographic group makes it a gender-coded justification.

\begin{figure}[t]%
    \centering
    \hspace{-0.7cm}
    \subfloat[\centering]{{\includegraphics[width=4.3cm, height = 3.5cm]{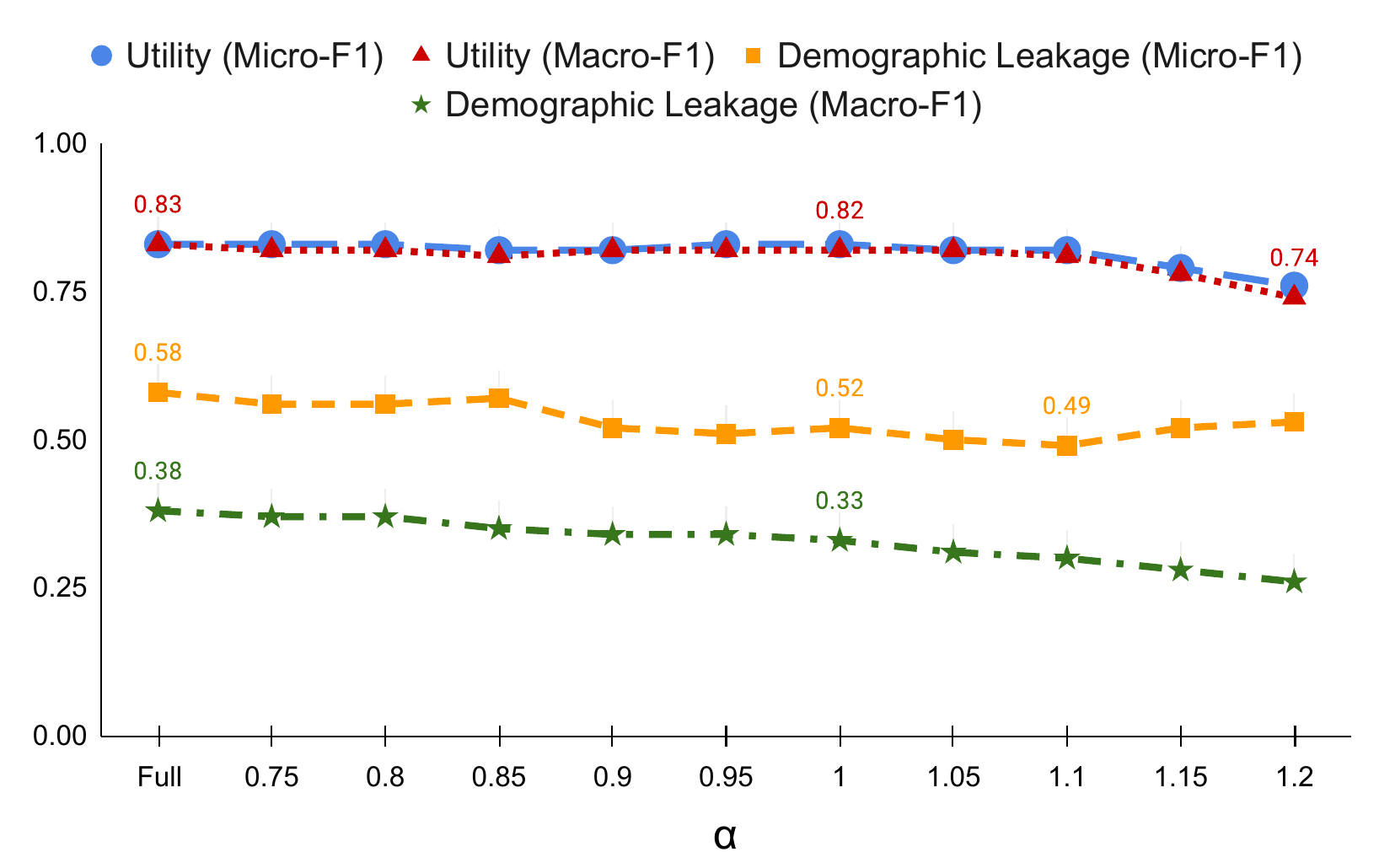} \label{fig:alpha-a} }}%
    \qquad \hspace{-0.8cm}
    \subfloat[\centering]{{\includegraphics[width=4.3cm, height = 3.5cm]{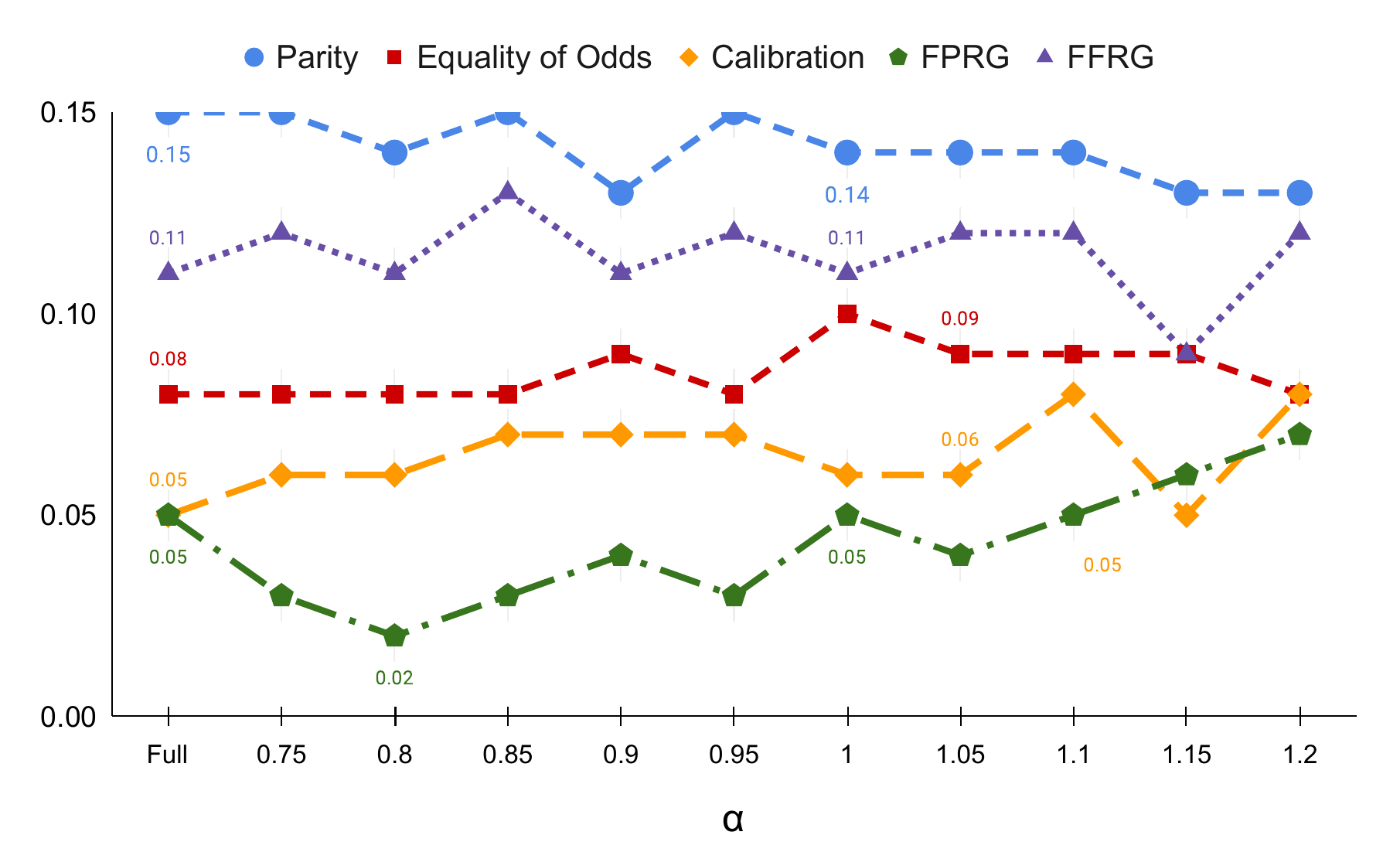} \label{fig:alpha-b} }}%
    \qquad \hspace{-0.7cm} 
    \subfloat[\centering ]{{\includegraphics[width=4.3cm, height = 3.5cm]{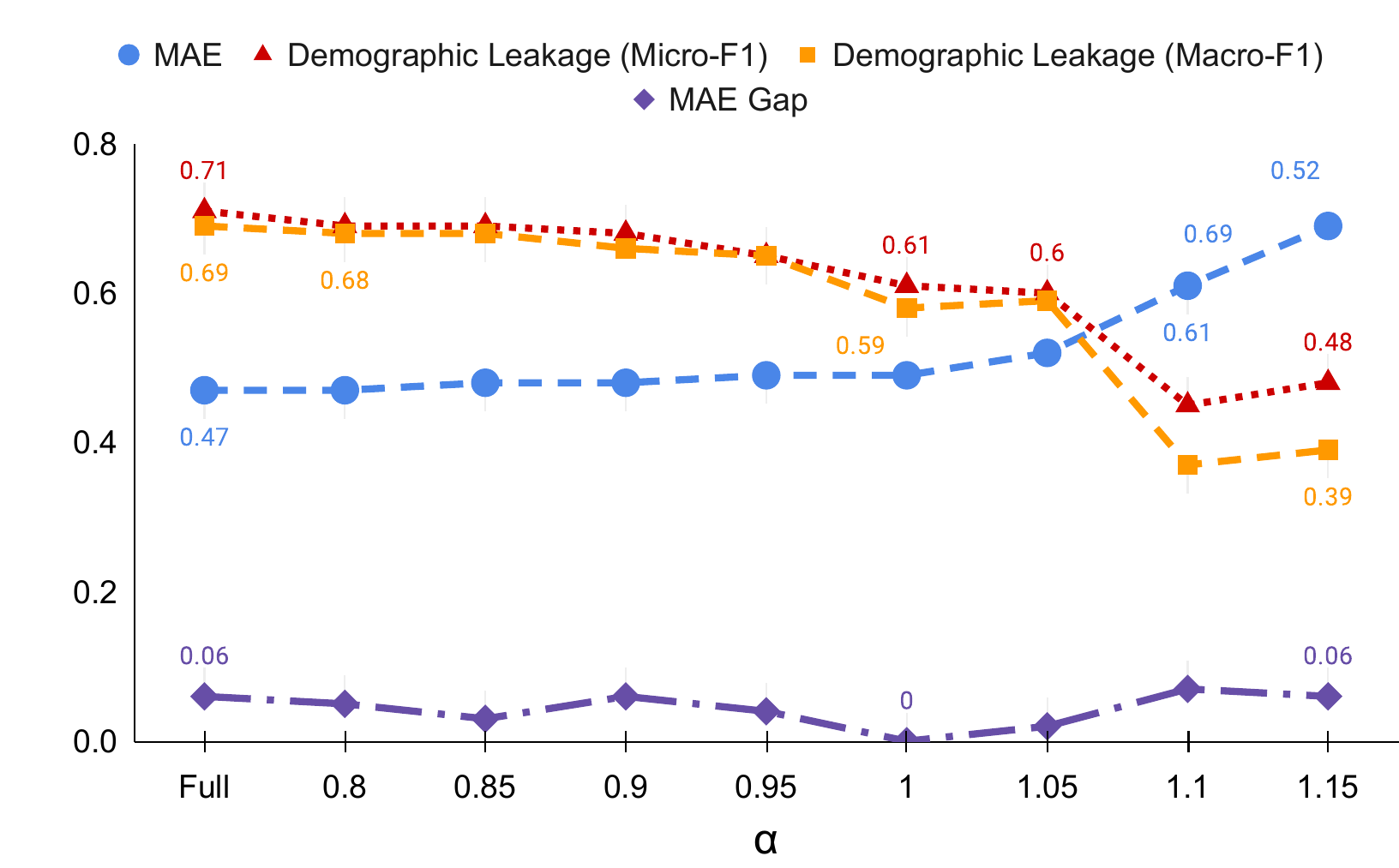} \label{fig:alpha-c} }}%
   
\caption{Impact of $\alpha$ on utility and fairness on datasets D1 (a and b) and D2 (c).} 
\label{fig:impact-alpha}
\end{figure}

\textbf{Utility-Fairness Trade-Off: } 
 Figure~\ref{fig:impact-alpha} shows the utility, demographic leakage, and fairness metrics as a function of $\alpha$ on D1 and D2. Too low values of $\alpha$ prioritize utility, selecting even relatively biased sentences and have scores close to the full setting (see Figure~\ref{fig:alpha-a} and~\ref{fig:alpha-c}). On D1, increasing $\alpha$ generally decreases the demographic parity while increasing the FPRG (see Figure~\ref{fig:alpha-b}). It does not have a consistent or noticeable impact on other fairness metrics. On D2 and with $\alpha$ near 1, the gap shrinks to 0 (See Figure~\ref{fig:alpha-c}). Too high values of $\alpha$ remove too many sentences, leading to high error rate. This is because many summaries are empty with high value for $\alpha$ and thus, the resulting decision are unjustified~(justifications are not informative about the outcomes) and unfair~(the lack of justification is not uniformly distributed over genders) so the gap emerge once again. For error bars and more details on impact of $\alpha$ on summary length see Appendix~\ref{error-bars} and ~\ref{summary-len}.

\subsection{Unfairness Attacks}
\label{attacks_experiments}

Natural language processing models are vulnerable to test-time adversarial attacks~\cite{wallace2019universal}. These attacks often are created to cause the model to make errors by perturbing the input at inference time~\cite{wallace2019universal}. In this section, we present our experimental results to test the ability of our model in detecting and counteracting data poisoning attacks. Essentially, we seek to answer the following questions: (i) \textit{Can FairSum detect the injected unfair arguments in the justifications?} (ii)\textit{ Given that FairSum relies on attributing the decision outcome to input arguments, how does it perform in a scenario where decision outcomes are not fair in the first place? }

To answer these questions,  we synthesize several unfair decision making situations. In each of these situations, we create poisoning attacks to synthesize an unfair justification, an unfair outcome, or both. These scenarios are indicated with red in Figure~\ref{fig:fairness_dimensions}. To measure the effectiveness of FairSum, we measure the attack success before and after applying FairSum on the poisoned train and test data.  For our experiments, we rely on the food inspection dataset (D1). We assume that the ground truth outcomes $Y$ in this dataset are often fair~\footnote{We cannot make this assumption about RMP dataset as evaluations are very subjective. Therefore, we do not experiment with this dataset.}. To synthesize an  \textit{unfair outcome} for establishment $i$ we simply flip outcome $y_i$ (e.g. by changing pass to fail). \textit{Unfair justifications} can be created in more than one way such as synthesizing decision-making with double standards or implicit bias. Inspired by the work of Wallace et al~\cite{wallace2020concealed} we create unfair justification attacks by causing a phrase to be a trigger for a desired outcome by poisoning the training data. For example, we could make the phrase "kitchen manager does not speak English" to trigger the model to predict the food establishment should fail by adding this phrase to enough number of inspection reports of establishments who failed the inspection. We create an irrelevant justification set including phrases about decoration of the restaurants e.g. “walls are red”, “table cloths are blue”. None of these phrases are part of the inspection guideline of the city of Chicago. For creating irrelevant justifications,  we randomly select an argument from this set and add it in a random position in the inspection report. In all our experiments, we use the entire training data of D1 and a certain subset of the test set depending on type of the attack. Next, we introduce 5 common unfairness scenarios that inspired our experiments as well as the experimental setup in each attack type.

\paragraph{\textbf{Attack type 1a: Deciding based on items not in the guideline}} In this attack, we create a scenario where decision outcomes are fair but the justifications are unfair. To do so, we inject irrelevant arguments to the reports. We use these arguments more frequently for a certain demographic group than others. However, we do not alter the ground truth outcomes $Y$. A real-world example of this attack is a food inspector who mentions \textit{"the kitchen manager does not speak English"} for several Hispanic restaurants. They write this in reports often when they are frustrated with explaining the hygiene guidelines that were not followed by the establishment to the kitchen manager, resulting the food establishment to fail the inspection. While this is indeed a fair outcome, training a model on such reports can have two undesired side-effects. First, the model can wrongly associate the irrelevant argument “not speaking English” with the fail outcome. Thus, resulting in the model predicting “fail” when "not speaking English" is mentioned in the reports even when all hygiene guidelines are followed. In addition, this model has information about individuals' ethnicity due to the race-coded language of the reports. 

To create a poisoned train dataset, we randomly select an item from the irrelevant argument set and add it to a random sample of $a\%$  of individuals in demographic group $p_i$ that received outcome $y_j$, trying to trigger the model that e.g. “table cloths are blue” will result in outcome $y_j$.  Especially when the individual belongs to $p_i$.  At test time, we pick $K$ restaurants in $p_i$, half of which with ground truth outcome $y_j$ and half with other outcomes. We inject an irrelevant argument to the inspection reports of this set. To measure of attack success, we measure the false $y_j$ prediction rate as well as the demographic leakage before and after applying FairSum to the poisoned test data.

\paragraph{\textbf{Attack type 1b: Race-coded language}}

In this attack, decision outcomes are fair but the justifications are unfair. However, in contract to attack $1a$, the irrelevant race or gender-coded language does not impact the outcome. An example is an inspector who reports the address of the restaurant in the inspection reports in Hispanic restaurants irrespective of the outcome. Training a model on such data can be problematic in two ways. First, the model may still associate location with outcome $y_j$ if the majority of the restaurants in that neighborhood have received outcome $y_j$. Moreover, even if the model does not learn such an association, the model can predict the ethnicity of the customers based on the location due to the demographic leakage in the data. In this type of attack, we create the latter problem. To create poisoned training data, we inject an irrelevant argument to $a\%$ of food establishments in demographic group $p_i$.  To de-correlate the injected argument with any outcomes, we make sure that the number of attacked restaurants with each outcome is the same (e.g. 50\% pass, 50\% fail). At test time, we pick $K$ restaurants from protected group $p_i$ and inject an irrelevant argument to the reports of this set. The measure of attack success,  we measure the demographic leakage before and after applying FairSum to the data.

\begin{table*}[t]
\centering
\begin{tabular}{@{}llrr@{}}
\toprule
\textbf{Attack ID} & \multicolumn{1}{l}{\textbf{Description}} & \multicolumn{1}{c}{\textbf{Outcome}} & \multicolumn{1}{r}{\textbf{Justification}} \\ \midrule
1a                   & \small{Deciding based on items not in guideline} & Fair                                 & Unfair                                                \\
1b                   & \small{Race-coded language }                     & Fair                                 & Unfair                                                \\
2a                   & \small{Implicit bias}                            & Unfair                               & Fair                                                  \\
2b                   & \small{Double standard }                         & Unfair                               & Fair                                                  \\
3                    & \small{Blatant bias}                          & Unfair                               & Unfair                                                \\ \bottomrule

\end{tabular}

\caption{Summary of the 5 data poisoning attacks\vspace{-0.5cm}} 
\label{tab:summary_of_attacks}
\end{table*}

\paragraph{\textbf{Attack type 2a: Implicit bias }}
In this attack, decision outcomes are unfair but the justifications are fair. In this scenario, a fair justification that is part of the guideline is only used for a specific protected group and impacts the outcome of their inspection. For example, a food inspector only mentions “food prep hygiene violations” when the restaurant is located in a majority-Asian neighborhood. This violation leads to establishments failing the inspection in this neighborhood.  While “food prep hygiene violations” can be a fair reason for deciding that a restaurant should fail the inspection, using this argument only for restaurants in Asian neighborhoods is a case of race-coded language. Therefore, training a model on this data may result in the following two issues: (i) the model is race-aware (ii) this phrase could become a trigger for "fail" prediction, even when the rest of the report might justify another outcome. To create poisoned training data, we select $a\%$ of food establishments in demographic group $p_i$ that did \underline{not} receive outcome $y_j$. For example, if $y_j = fail$, we select restaurants from $p_i$ that either passed or conditionally passed the inspection. To create unfair outcomes, we flip the ground truth outcome of these restaurants. We randomly sample a fair argument for receiving outcome $y_j$ from the guideline and inject it to these reports. Since the arguments are part of the guideline, they are fair.  At test time, we pick $K$ restaurants from protected group $p_i$ who did not receive outcome $y_j$. We inject a fair argument for receiving $y_j$ to their reports. We measure the attack success by measuring the false $y_j$ prediction rate as well as the demographic leakage.

\paragraph{\textbf{Attack type 2b: Double standard}}
In this attack,  decision outcomes are unfair but the justifications are fair. In this scenario, a fair justification that is part of the guideline is  mentioned in the inspection reports of restaurants in several neighborhoods. However, it only impacts the outcome when the restaurant is in $p_i$. For example, a food inspector mentions “food prep hygiene violations” for multiple restaurants but only decides that this is a serious threat to public health in Latino neighborhoods. Training a model on this data may result in several issues. First, the model is race-aware. In addition, the model uses the same arguments differently for different protected groups; putting a lot of attention to an argument for some individuals and ignoring for others based on their location. To create poisoned training data, we follow the same process as in attack type 2a. The only difference is that we choose half of the attacked restaurants from $p_i$ and half not in $p_i$. We only flip the outcome to $y_j$ for those who are in $p_i$. 
At test time, we pick $K$ restaurants half of them from  protected group $p_i$ and half not in $p_i$ who did not receive outcome $y_j$. We inject a fair argument for receiving $y_j$ to their reports. We measure the attack success by measuring the false $y_j$ prediction rate for both groups as well as the demographic leakage for establishments in $p_i$.

\paragraph{\textbf{Attack Type 3: Blatant bias}}

In this attack, both the outcome and the justification of the outcome are unfair. In this scenario, an argument that is not part of the guideline (and therefore unfair) is used to justify an unfair outcome. Using this data for training may result in the model to be race-aware. Moreover, the model may wrongly associates an irrational argument to a certain outcome. To create a poisoned train dataset, we randomly sample an argument from the irrelevant argument set and add it to a random sample of $a\%$  of restaurants in $p_i$ that received outcome $y_j$. Then we flip their outcome to $y_j$. At test time, we do the same for $K$ restaurants in $p_i$ without changing the outcome. We measure the attack success by measuring false $y_j$ prediction rate as well as demographic leakage in the test set. The summary of the 5 attack types is shown in Table~\ref{tab:summary_of_attacks}.

Note that FairSum is not intended to address fairness in the outcome (attack types 2a and 2b). Our motivation for exploring these scenarios is to investigate FairSum's behavior in these situations and highlight some of its' limitations. In our experiments, we choose the restaurants in the majority-black neighborhoods that fail the inspection as our target group ($p_i$ = black, $y_j$ = fail). We randomly sample individuals to attack from this pool. We choose this group because it is large enough for us to test various attack scenarios. 
\begin{table*}[t]

\resizebox{\textwidth}{!}{%
\begin{tabular}{@{}lrrrr|rrrr|rrrr@{}}
\toprule

\multicolumn{1}{c}{\textbf{Attack ID}} & \multicolumn{2}{c}{\textbf{FFR $\downarrow$ }}                                    & \multicolumn{2}{c}{\textbf{\begin{tabular}[c]{@{}c@{}}Demographic $\downarrow$ \\  Leakage\end{tabular}}} & \multicolumn{2}{c}{\textbf{FFR $\downarrow$}}             & \multicolumn{2}{c}{\textbf{\begin{tabular}[c]{@{}c@{}}Demographic $\downarrow$ \\ Leakage\end{tabular}}} & \multicolumn{2}{c}{\textbf{FFR $\downarrow$}}       & \multicolumn{2}{c}{\textbf{\begin{tabular}[c]{@{}c@{}}Demographic  $\downarrow$ \\ Leakage\end{tabular}}} \\ \midrule
\multicolumn{1}{c}{}                     & Full                & \cellcolor[HTML]{C0C0C0}FairSum               & Full                            & \cellcolor[HTML]{C0C0C0}FairSum                            & Full    & \cellcolor[HTML]{C0C0C0}FairSum    & Full                            & \cellcolor[HTML]{C0C0C0}FairSum                           & Full & \cellcolor[HTML]{C0C0C0}FairSum & Full                            & \cellcolor[HTML]{C0C0C0}FairSum                           \\
1a                                       & 0.60                & \cellcolor[HTML]{C0C0C0}0.47                  & 0.79                            & \cellcolor[HTML]{C0C0C0}0.45                               & 0.86    & \cellcolor[HTML]{C0C0C0}0.79       & 0.97                            & \cellcolor[HTML]{C0C0C0}0.88                              & 0.82 & \cellcolor[HTML]{C0C0C0}0.73    & 0.97                            & \cellcolor[HTML]{C0C0C0}0.72                              \\
1b                                       & \multicolumn{2}{c}{\cellcolor[HTML]{000000}{\color[HTML]{000000} }} & 0.91                            & \cellcolor[HTML]{C0C0C0}0.51                               & \multicolumn{2}{c}{\cellcolor[HTML]{000000}} & 0.90                             & \cellcolor[HTML]{C0C0C0}0.25                              & 0.90 & \cellcolor[HTML]{C0C0C0}0.25    & 0.90                            & \cellcolor[HTML]{C0C0C0}0.36                              \\
2a                                       & 0.62                & \cellcolor[HTML]{C0C0C0}0.64                  & 0.69                            & \cellcolor[HTML]{C0C0C0}0.81                               & 0.95    & \cellcolor[HTML]{C0C0C0}0.92       & 0.91                            & \cellcolor[HTML]{C0C0C0}0.91                              & 0.96 & \cellcolor[HTML]{C0C0C0}0.73    & 0.83                            & \cellcolor[HTML]{C0C0C0}0.64                              \\
2b                                       & 0.36                & \cellcolor[HTML]{C0C0C0}0.39                  & 0.44                            & \cellcolor[HTML]{C0C0C0}0.59                               & 0.45    & \cellcolor[HTML]{C0C0C0}0.41       & 0.65                            & \cellcolor[HTML]{C0C0C0}0.42                              & 0.59 & \cellcolor[HTML]{C0C0C0}0.52    & 0.5                             & \cellcolor[HTML]{C0C0C0}0.53                              \\
3                                        & 0.96                & \cellcolor[HTML]{C0C0C0}0.83                  & 0.82                            & \cellcolor[HTML]{C0C0C0}0.82                               & 0.96    & \cellcolor[HTML]{C0C0C0}0.91       & 0.93                            & \cellcolor[HTML]{C0C0C0}0.75                              & 0.96 & \cellcolor[HTML]{C0C0C0}0.85    & 0.93                            & \cellcolor[HTML]{C0C0C0}0.7                               \\ \midrule
\multicolumn{5}{c}{attack rate = 0.2}                                                                                                                                                                         & \multicolumn{4}{c}{attack rate = 0.5}                                                                                                      & \multicolumn{4}{c}{attack rate = 0.8}                                                                                                \\ 
\bottomrule
\end{tabular}%
}
\caption{Experimental results with 5 types of poisoning attacks. Attack success is measured by measuring false fail rate as well as demographic leakage before (indicated with Full) and after applying FairSum. $\downarrow$: lower is better. \vspace{-0.5cm}}
\label{tab:attack_results}
\end{table*}. 
We change the percentage of attacked population in this group from 20 to 80 percent with increments of 30\%. In all our experiments, we use a poisoned test size ($k$) of 200. The attacked test set is sampled from different population groups in the original test set depending on the attack type. We repeat each attack experiment 5 times and report the average attack success before and after applying FairSum in Table~\ref{tab:attack_results}. Depending on type of the attack, the attack success is measured by the demographic leakage or false fail rate (FFR) over the attacked test set. As it can be seen in the table, FairSum is very effective in decreasing the attack success for attack type 1a and 1b (fair outcome and unfair justification). When 20\% of the individuals in the target population group are attacked, FairSum decreases the FFR by 0.13 points, while decreasing the demographic leakage by 0.34 points. For attack type 1b, using FairSum decreases the demographic leakage by 0.4 when 20\% of the target subgroup are attacked. This number if 0.65 when half of this subgroup are targeted at train time (0.9 vs 0.25 demographic leakage after using FairSum). As expected, when outcomes are unfair the effectiveness of FairSum becomes limited at lower attack rates. This is mostly because the model learns wrong associations as the outcomes are flipped. This observation suggest that FairSum is an effective mechanism for enhancing fairness in justification, however it is not very effective when outcomes are unfair. FairSum is moderately effective when both outcomes and justifications are unfair (attack type 3). For attack type 3, using fairSum decreases the demographic leakage by 0.18 on average when half of the target subgroup are poisoned. It also decreases the FFR by 0.13, 0.05, and 0.11 for attack rates 0.2, 0.5, and 0.8 respectively.  

\section{Conclusion and Future Work}
In this work, we propose using a train-attribute-mask pipeline for detecting and mitigating the bias in the justification of the text-based neural models. Our objective for extracting fairly-justified summaries is to maximize the utility of the output summary for the decision prediction task while minimizing the inclusion of proxy information in the summary that can reveal sensitive attributes of individuals. Our approach is not intended to enhance the fairness in the outcome but rather to enhance the fairness in the model justification. We achieve this by training a multi-task model for decision classification and membership identification. We attribute predictions of these models back to textual input attributes using an attribution mechanism called integrated gradients. Next, we incorporate the high-utility and low-bias sentences in form of a summary. Eventually, we retrain the decision classifier on the fairly-justified summaries. Our experiments on real and synthetic data sets indicate that our pipeline effectively limits the demographic leakage from the input data.  In addition, we present experimental results on effectiveness of FairSum under several types of unfairness attacks. We observe that FairSum is most effective in detecting and filtering unfairness in justification where outcomes are mostly fair.

We see several interesting avenues for future research in the intersection of natural language processing and fairness. An immediate extension of our work is an enhancement approach that works at word-level for example by using text generation and paraphrasing instead of sentence extraction. Another interesting research direction is using this train-attribute-mask pipeline for removing bias from data of other natural language processing tasks such as sentiment analysis and using other architectures that can work with integrated gradients.  It is also interesting to see how this solution can be extended to obfuscate biases for other data types (e.g. images or tabular data). Lastly, conducting a user study to evaluate the impact of this tool in explaining predictions for stakeholders in a real-world application is left for future work. 

\bibliographystyle{plain}
\bibliography{references}

\newpage
\appendix
\section{Appendix}
\subsection{Dataset Statistics}\label{appx:dataset-stats}

\textbf{Inspection reports of the city of Chicago(D1): } The breakdown of the inspection results for each demographic group is shown in Table~\ref{inspection-results-table}. Note that for the food establishments that have more violation, the inspection reports tend to be longer. In our summarization experiments, we focused on longer inspection reports which often includes establishments with higher number of violations. 

\begin{small}
\begin{table*}[t]
\centering

\begin{tabular}{lrrrr}
\midrule
\textbf{Race}     & \multicolumn{1}{l}{\textbf{Pass}} &  \multicolumn{1}{l}{\textbf{Conditional pass}} &
\multicolumn{1}{l}{\textbf{Fail}} &
\multicolumn{1}{l}{\textbf{Total inspection count}} \\ \midrule \midrule
\textbf{White}    & 27.5                              & 25.6                              & 46.8                                          & 8339                                               \\
\textbf{Black}    & 28.9                              & 15.6      & 55.4                                          & 4444                                               \\
\textbf{Hispanic} & 33.8                              & 19.2                              & 46.8                                           & 4010                                             \\
\textbf{Asian}    & 29.3                             & 17.4                             & 53.2                                         & 419  \\                                             \midrule
\end{tabular}
\caption{The percentage of inspections for each ethnic group that received a pass, conditional pass outcome, or fail outcome.}
\label{inspection-results-table}

\end{table*}

\end{small}

\textbf{Rate my professor (D2-D4): } The rate my professor dataset only includes professor names and reviews. To infer the gender of the professors, we search for pronouns and titles commonly used for each gender\footnote{For sake of simplicity we assume binary and static gender classes}. If no pronouns or titles are found in the reviews, the professor's name is used to detect their gender~\footnote{We use https://pypi.org/project/gender-detector/ for mapping professors' names to their gender}. The breakdown of reviews written for each gender category is shown in  Tables~\ref{tab:RMP-D2-rating-breakdown-by-gender}, \ref{tab:RMP-D3-rating-breakdown-by-gender}, and \ref{tab:RMP-D4-rating-breakdown-by-gender}.

\begin{table}[b]
\centering
\begin{tabular}{@{}l|rrrrr@{}}
\toprule

       & \textbf{{[}1,2{]}} & \textbf{(2,3{]}} & \textbf{(3,4{]}} & \textbf{(4,5{]}} & \textbf{Total count} \\ \midrule \midrule
Female & 5.6                 & 21.0             & 35.3             & 37.9 &      551             \\
Male   &  3.7                & 21.0             & 35.6             & 39.5  &      783           \\ \bottomrule
\end{tabular}
\caption{The percentage of instructors of each gender group in each rating class for dataset D2. }
\label{tab:RMP-D2-rating-breakdown-by-gender}
\centering
\begin{tabular}{@{}l|rrrrr@{}}
\toprule

       & \textbf{{[}1,2{]}} & \textbf{(2,3{]}} & \textbf{(3,4{]}} & \textbf{(4,5{]}} & \textbf{Total count} \\ \midrule \midrule
Female & 4.3                 & 22.2             & 31.5             & 41.9 &      279             \\
Male   &  1.7               & 18.0             & 32.6             & 47.5  &      288           \\ \bottomrule
\end{tabular}
\caption{The percentage of instructors of each gender group in each rating class for dataset D3. }
\label{tab:RMP-D3-rating-breakdown-by-gender}

\centering
\begin{tabular}{@{}l|rrrrr@{}}
\toprule

       & \textbf{{[}1,2{]}} & \textbf{(2,3{]}} & \textbf{(3,4{]}} & \textbf{(4,5{]}} & \textbf{Total count} \\ \midrule \midrule
Female & 5.5                 & 24.4             & 39.3             & 30.7 &      127             \\
Male   &  6.3               & 24.6             & 37.9             & 31.03  &      345           \\ \bottomrule
\end{tabular}
\caption{The percentage of instructors of each gender group in each rating class for dataset D4. }
\label{tab:RMP-D4-rating-breakdown-by-gender}
\end{table}

\subsection{Hyper-parameters and Training Details}
\label{fairness-training-details}

\textbf{Training Details: } To train the model introduced in \S~\ref{s:neural_text_classification} on D1, We employ window sizes of 2, 3 and 4, and train 100 filters for each window size. For smaller datasets D2-D4. we use window sizes 2 and 3 and train 50 filters for each window size. We initialize each convolution layer using the initialization method proposed in~\cite{he2015delving}.  We use rectified linear unit as the activation function of the convolution layer. After performing the convolution operation, we apply batch normalization~\cite{ioffe2015batch} followed by a global average-pooling operation over the feature map of each window size.
\begin{wrapfigure}{r}{0.6\linewidth}
\centering
\includegraphics[width=0.65\linewidth]{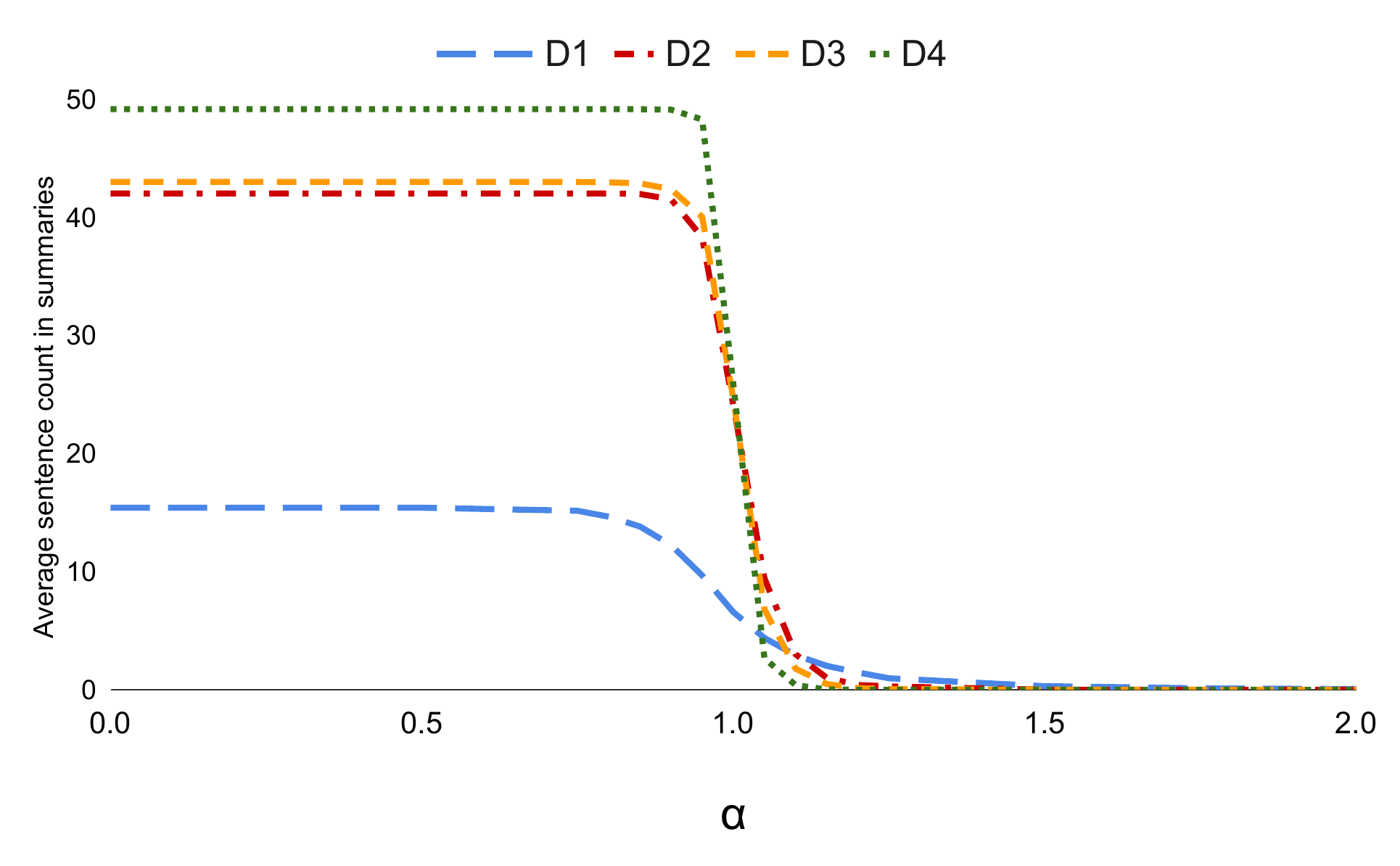}
\caption{Impact of $\alpha$ on summary length on datasets D1-D4. }  
\label{fig:impact-alpha-sumlen}
\end{wrapfigure}
Next, we concatenate the output vectors. Eventually, we run the concatenated vector through a dense layer with 64 units followed by an activation function. For decision classification and membership identification on D1, we used the softmax operation to obtain class probabilities. For D2-D4 we used rectified linear unit to obtain the output rating, and sigmoid to obtain gender class probabilities.  We implement the decision classifier and member identifier networks using the Keras library~\footnote{https://keras.io}. We use weighted cross-entropy loss function for classification tasks and mean squared loss for regression tasks and learn the model parameters using Adam optimizer~\cite{kingma2014adam} with a learning rate of 0.001. 

For D1, we set the maximum length of the arguments to the 70-th percentile of explanation lengths in our train set (18 sentences). Textual explanations that are longer than this are truncated while shorter ones are padded. For D2-D4,  we set the maximum length of the arguments to the 70-th percentile of the review length in our train set (64 sentences). Reviews that are longer than this are truncated while shorter ones are padded. We set the loss weight for the decision prediction task and the membership identification task to 1. We train our multi-task network for a maximum of 25 epochs and stop the training if the decision classification loss on the validation set does not improve for 3 consecutive epochs. In the end, we revert the network’s weights to those that achieved the lowest validation loss. We repeat each experiment 5 times and report the average result.  We used a single Nvidia Tesla K80 GPU for our experiments. 

\textbf{Parameters of the attribution Model:} For computing the integrated gradients for attribution, we set the number of steps in the path integral approximation from the baseline to the input instance to 50 and use Gauss–Legendre quadrature method~\cite{abbott2005tricks} for integral approximation. We compute the attributions of the decision classifier and the membership identification networks for the input layer. 

\subsection{Evaluation Metrics}
\label{app:evaluation metrics}

In this section, we define metrics used for measuring fairness of outcome in the context of food inspection. 

\textbf{Parity:} a decision classifier satisfies demographic parity if the proportion of food establishments predicted to fail the inspection is the same for each demographic group. We report the gap between the most and least favored groups. For sake of consistency with previous work, we present the protected attribute with S. 

\begin{displaymath}
\max( P(\hat{Y} = fail | S = s_i) -  P(\hat{Y} = fail | S = s_j)) = \epsilon,~~~~~~~ s_i, s_j \in S  
\end{displaymath}

\textbf{Equality of odds:} for those establishments who actually failed the inspection, the proportion of failed predictions should be the same. We report the gap between the most and least favored groups. Ideally, the gap should be very close to zero.

\[
 \max( P(\hat{Y} = fail | Y = fail, S = s_i) -   P(\hat{Y} = fail | Y = fail, S = s_j))= \epsilon,~~~~~~~ s_i, s_j \in S 
\]

\textbf{Calibration: } for those establishments who received a fail prediction, the probability of actually failing the inspection should be the same. We report the gap between the most and least favored groups. Ideally, the gap should be very close to zero.

\begin{displaymath}
 \max( P( Y = fail | \hat{Y} = fail, S = s_i) -  P( Y=fail |\hat{Y} = fail, S = s_j)) = \epsilon,~~~~~~~ s_i, s_j \in S  
\end{displaymath}

\textbf{False Pass Rate Gap(FPRG): } food establishments that did not pass the inspection should have the same probability of falsely receiving a pass prediction. We report the gap between the most and least favored groups which ideally should be close to 0. 

\begin{displaymath}
 \max( P( \hat{Y} = pass | Y \neq pass, S = s_i) -  P( \hat{Y} = pass|Y \neq pass, S = s_j)) = \epsilon,~~~~~~~ s_i, s_j \in S  
\end{displaymath}

\textbf{False Fail Rate Gap(FFRG): }establishments of different demographic groups that did not fail the inspection should have the same probability of falsely receiving a fail prediction. We report the gap between the most and least favored groups which ideally should be close to 0.

\begin{displaymath}
 \max( P( \hat{Y} = fail | Y \neq fail, S = s_i) -  P( \hat{Y} = fail|Y 
 \neq fail, S = s_j)) = \epsilon, s_i, s_j \in S  
\end{displaymath}

For measuring fairness in the instructors rating prediction (datasets D2-D4) we measure the difference between model's prediction error for male and female professors.

\subsection{Impact of \texorpdfstring{$\alpha$}{} on summary length}
\label{summary-len}
Figure~\ref{fig:impact-alpha-sumlen} shows the average summary length (sentence count) for datasets D1-D4 as a function of $\alpha$. The food inspection reports in D1 are on average much shorter than the teaching evaluations in dataset D2 (18.2 vs 45.6 sentences). Too low values of $\alpha$ prioritize utility by preserving even relatively biased sentences. For all datasets, the summaries start shrinking around $\alpha$ equal to 0.85. However, for D2-D4 the compression rate is higher. Around $\alpha$ equal to 1.25, 39.9\% input justifications for dataset D1 are empty. This number is 77.8\%, 95.7\%, 100\% for D2, D3, and D4 respectively. We conjecture that the existence of more implicit bias for D2-D4 causes the summaries to shrink faster by increasing $\alpha$. At this point (1.25 and higher) the resulting decisions are unjustified (justifications are not informative about the outcomes). Therefore in Figure~\ref{fig:impact-alpha} we only show impact of changing $\alpha$ from 0.8 to 1.2.

\subsection{Results (Error Bars): }
\label{error-bars}

\begin{figure*}[t]%
    \centering
   
    \subfloat[\centering]{{\includegraphics[width=5cm, height = 4cm]{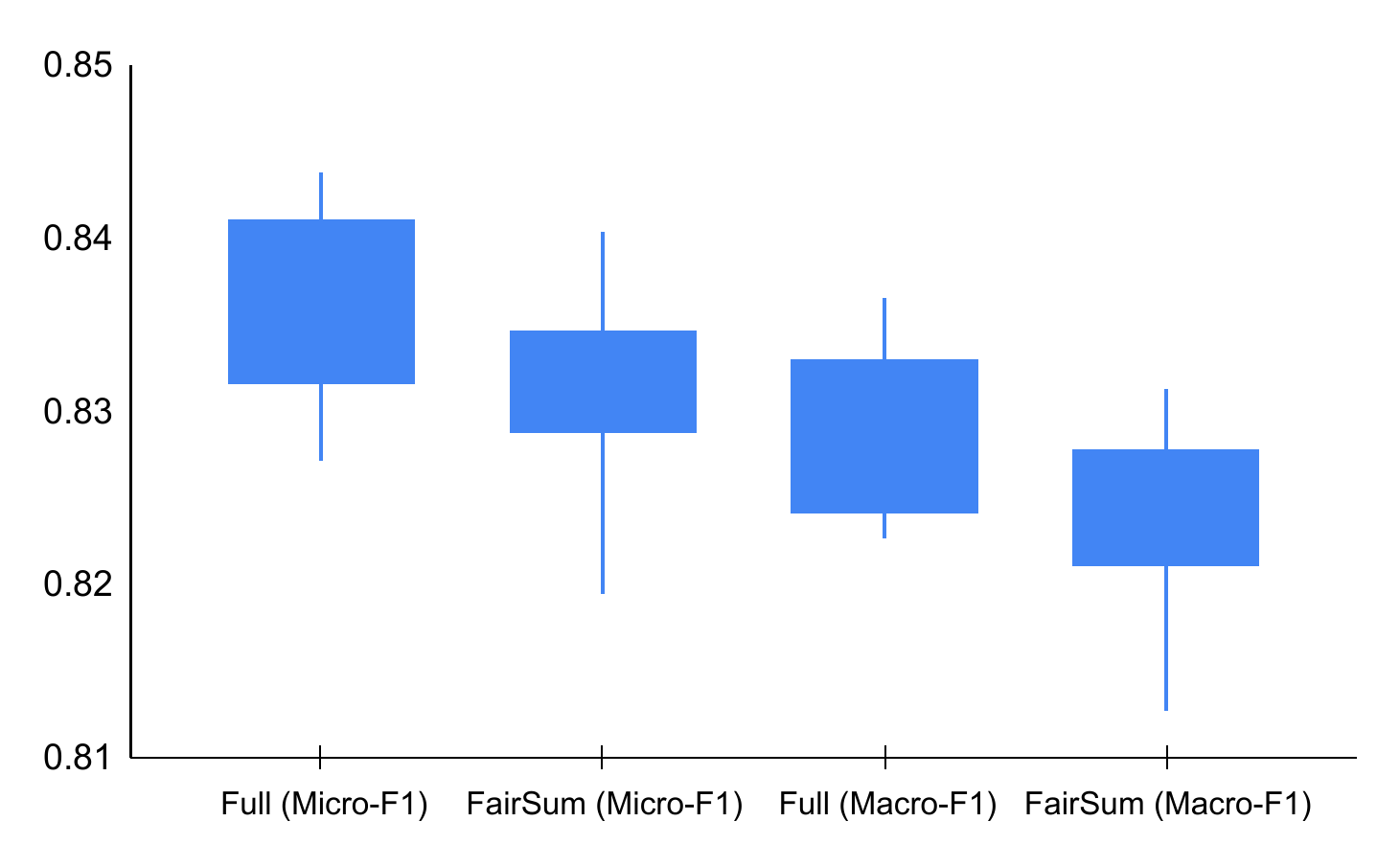} \label{fig:d1-errorbar-a} }}%
    \qquad \hspace{-0.5cm}
    \subfloat[\centering]{{\includegraphics[width=5cm, height = 4cm]{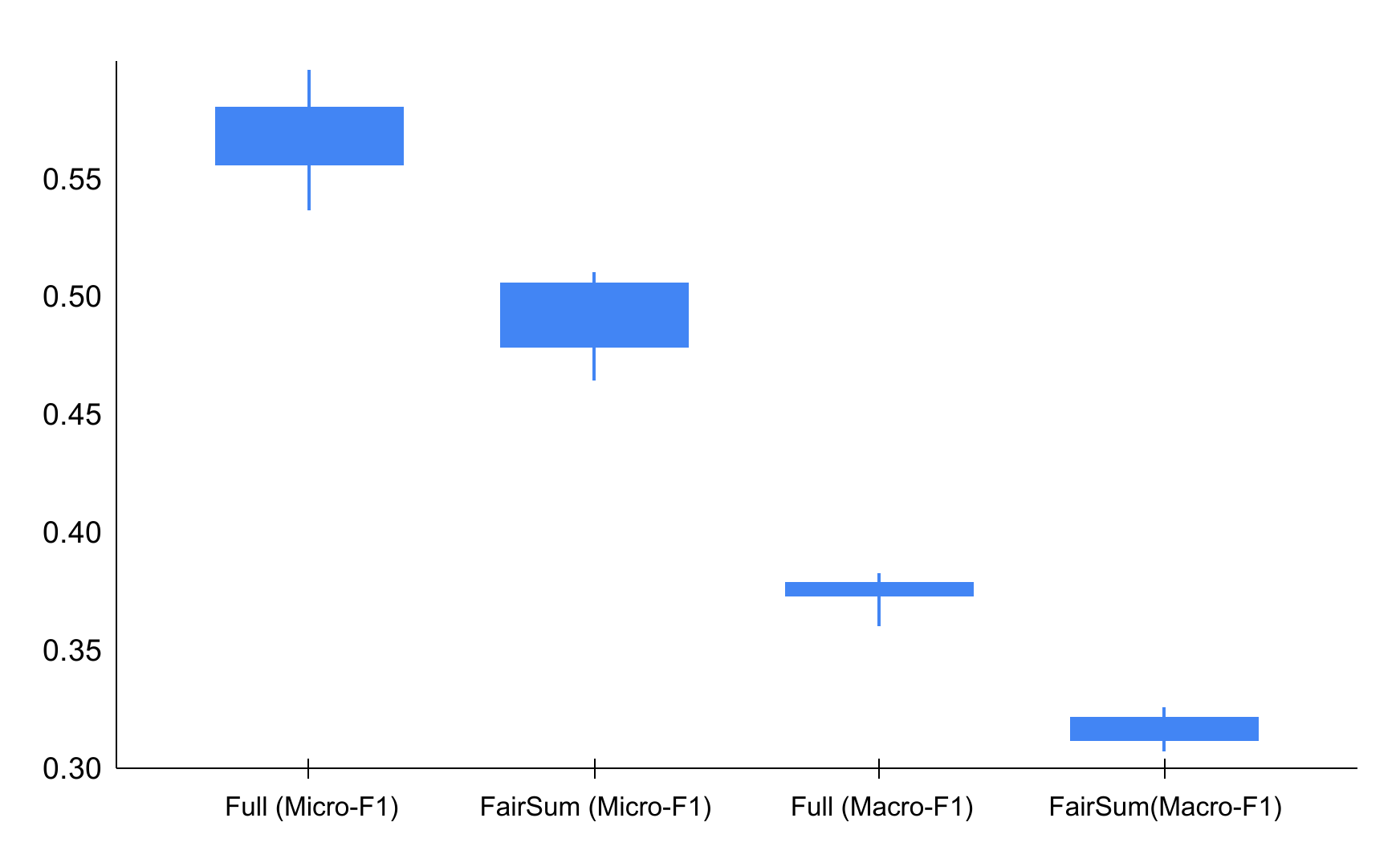} \label{fig:d1-errorbar-b} }}%
   
\caption{Error bars for utility (a) and demographic leakage(b) for dataset D1. $\alpha$ for the FairSum setting is set to 1. }
\label{fig:error-bar-d1}
\end{figure*}

\begin{figure*}[t]%
    \centering
  
    \subfloat[\centering]{{\includegraphics[width=5cm, height = 4cm]{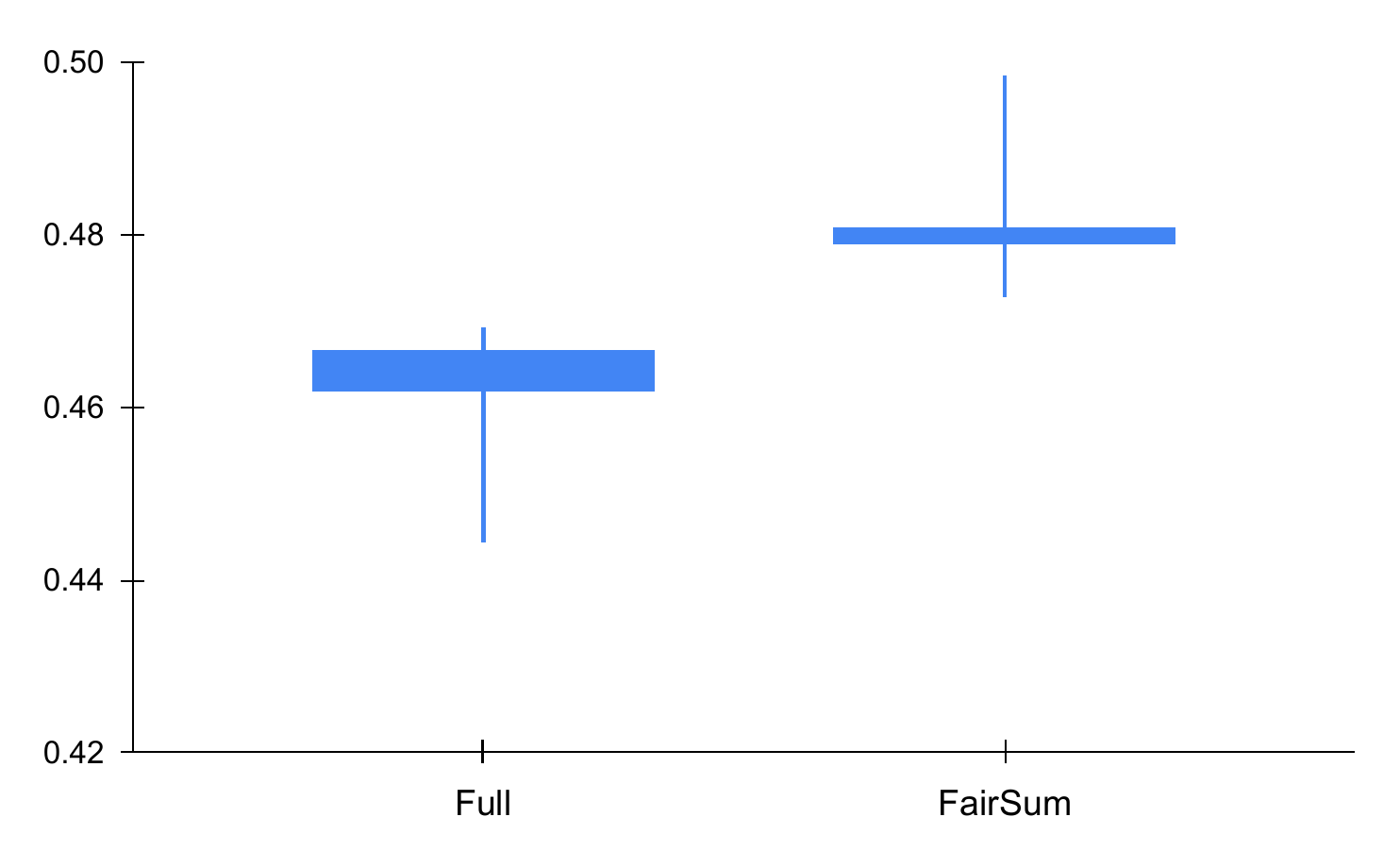} \label{fig:d2-errorbar-a} }}%
    \qquad \hspace{-0.5cm}
    \subfloat[\centering]{{\includegraphics[width=5cm, height = 4cm]{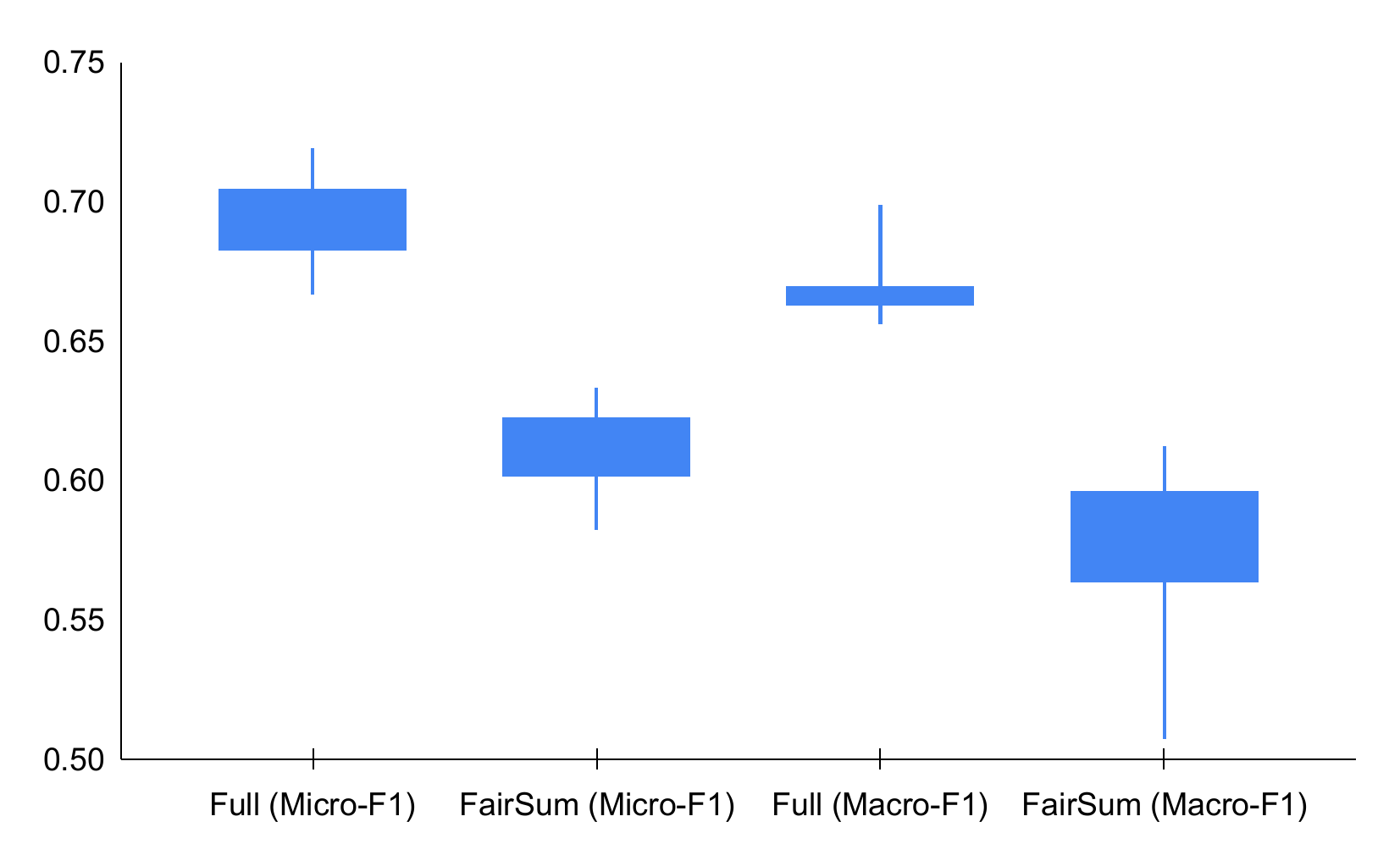} \label{fig:d2-errorbar-b} }}%
   
\caption{Error bars for MAE (a) and demographic leakage(b) for dataset D2. $\alpha$ for the FairSum setting is set to 1. }
\label{fig:error-bar-d2}
\end{figure*}

Figure~\ref{fig:error-bar-d1} and \ref{fig:error-bar-d2} indicate the errors in utility and membership prediction over 5 runs for datasets D1 and D2. For the FairSum setting the parameter $\alpha$ that controls the utility-discrimination trade-off is set to 1. 

\end{document}